\documentclass[11pt, a4paper, logo, copyright]{googledeepmind}

\usepackage[authoryear, sort&compress, round]{natbib}
\usepackage{multicol}
\usepackage{multirow}
\usepackage{subcaption}
\usepackage{tablefootnote}
\usepackage{listings}
\usepackage{pifont}

\usepackage{float}

\usepackage{amsmath, amssymb}
\usepackage{xcolor}
\usepackage{mdframed} % For framing quotes and logs
\usepackage{enumitem} % For better list formatting

\usepackage{hyperref}
\usepackage{tabularx}
\usepackage{booktabs}
\usepackage[table]{xcolor}
\usepackage{ragged2e}
\usepackage[most]{tcolorbox}

\defcitealias{Feng2026}{Feng26}
\defcitealias{FYZ4}{FYZ26}
\defcitealias{FK26}{FK26}
\defcitealias{LeeSeo2026}{LeeSeo26}
\defcitealias{ACGKMP}{ACGKMP26}
\defcitealias{BKKKZ}{BKKKZ26}

\defcitealias{schmitt2025extremal}{Schmitt25}
\defcitealias{bryan2026motivic}{BEMSV26}
\defcitealias{jang2025point}{JangRyu25}
\defcitealias{ivanisvili2025counterexamplemajorityoptimalitynicd}{IvanisviliXie25}
\defcitealias{salim2025acceleratingmathematicalresearchlanguage}{Salim25}
\defcitealias{alexeev2025forbidden}{AlexeevMixon25}
\defcitealias{feldman2025godeltestlargelanguage}{FeldmanKarbasi25}
\defcitealias{georgiev2025mathematicalexplorationdiscoveryscale}{GGTW25}
\defcitealias{Hir58}{Hir58}
\defcitealias{Gro97}{Gro97}
\defcitealias{Mum77}{Mum77}
\defcitealias{GL14}{GL14}
\defcitealias{sah2019number}{SSSZ19}
\defcitealias{Bannai81}{BB81}

\definecolor{humanbubble}{RGB}{235, 245, 255}
\definecolor{aibubble}{RGB}{245, 245, 245}
\definecolor{loggray}{RGB}{230, 230, 230}
\definecolor{indexbg}{RGB}{248, 248, 248}

\newtcolorbox{interactionlog}[2][]{
  enhanced,
  arc=0pt, outer arc=0pt,
  colback=white, colframe=black!60,
  boxrule=0.8pt,
  fonttitle=\bfseries\sffamily, coltitle=black, colbacktitle=loggray,
  title={Human-AI Interaction Card \if\relax\detokenize{#1}\relax\else for #1\fi},
  halign title=center, attach title to upper,
  after title={\vspace{4pt}\hrule\vspace{10pt}},
  lower separated=true,
  segmentation style={solid, black!60, line width=0.8pt},
  colbacklower=indexbg,
  after upper={\par\vfill
    \begin{tcolorbox}[
      enhanced, colback=indexbg, colframe=white, boxrule=0pt,
      top=0pt, bottom=0pt, fontupper=\footnotesize\sffamily,
      title=Raw prompts and outputs, coltitle=black!70, attach title to upper,
      after title={:\enskip}, sharp corners
    ]
    #2 
    \end{tcolorbox}
  }
}

\newcommand{\human}[1]{%
  \noindent\begin{flushright}
    \begin{minipage}[c]{0.70\textwidth}
      \begin{tcolorbox}[
        enhanced,
        colback=humanbubble, colframe=black!15,
        arc=6pt, sharp corners=southeast, boxrule=0.5pt,
        left=6pt, right=6pt, top=4pt, bottom=4pt, boxsep=0pt
      ]\small #1\end{tcolorbox}
    \end{minipage}%
    \hspace{8pt}
    \begin{minipage}[c]{40pt}
      \footnotesize\sffamily\textbf{Human}
    \end{minipage}
  \end{flushright}
  \vspace{-12pt}
}

\newcommand{\ai}[2]{%
  \noindent\begin{flushleft}
    \begin{minipage}[c]{55pt}
      \footnotesize\sffamily\textbf{#1}
    \end{minipage}%
    \hspace{2pt}
    \begin{minipage}[c]{0.65\textwidth}
      \begin{tcolorbox}[
        enhanced,
        colback=aibubble, colframe=black!15,
        arc=6pt, sharp corners=southwest, boxrule=0.5pt,
        left=6pt, right=6pt, top=4pt, bottom=4pt, boxsep=0pt
      ]\small #2\end{tcolorbox}
    \end{minipage}
  \end{flushleft}
  \vspace{-12pt}
}

\newtcolorbox{LLMBox}[2][]{
    colback=gray!5!white,
    colframe=gray!75!black,
    fonttitle=\bfseries,
    title=#2,               % The title is the second mandatory argument
    arc=2mm,
    boxrule=0.5pt,
    left=8pt,               % Slightly tighter margins for smaller boxes
    right=8pt,
    top=8pt,
    bottom=8pt,
    #1                      % This allows width=..., fontupper=..., etc.
}

\definecolor{tablegray}{HTML}{EFEFEF}

\renewcommand{\cite}[1]{\citep{#1}}

\newcommand{\erdos}[1]{Erd\H{o}s-{#1}}
\newcommand{\aletheia}{\emph{Aletheia}}
\newcommand{\fp}{\emph{FirstProof}}
\newcommand{\dtvt}{Gemini Deep Think (advanced version, Jan 2026)}
\newcommand{\ipb}{\textit{IMO-ProofBench}}
\newcommand{\fm}{\textit{FutureMath}}
\usepackage{soul}

\definecolor{ZSBaseline}{HTML}{ff8d13}
\definecolor{KDBaseline}{HTML}{bd00ff}
\definecolor{DABaseline}{HTML}{2782ed}
\definecolor{OurColor}{HTML}{36aa70}
\definecolor{ExampleBg}{HTML}{ffffff}
\definecolor{ExampleTitle}{HTML}{545f7f}
\newmdenv[
    roundcorner=5pt,
    backgroundcolor=ExampleBg,
    linecolor=ExampleTitle,
    outerlinewidth=0.5pt,
    frametitlebackgroundcolor=ExampleTitle,
    frametitlefont={\bfseries\color{white}},
    nobreak=true,  % remove this if you want to allow breaking across columns, but i thought it was pretty messy
]{problemexample}

\AtBeginEnvironment{problemexample}{\setlength{\parindent}{0pt}}

\title{Towards Autonomous Mathematics Research}

\author{
 Tony Feng\textsuperscript{*$\dagger$},
 Trieu H. Trinh,
 Garrett Bingham,
 Dawsen Hwang,
 Yuri Chervonyi,
 Junehyuk Jung\textsuperscript{$\dagger$},
 Joonkyung Lee\textsuperscript{$\dagger$},
 Carlo Pagano\textsuperscript{$\dagger$},
 Sang-hyun Kim\textsuperscript{$\dagger$},
 Federico Pasqualotto\textsuperscript{$\dagger$},
 Sergei Gukov\textsuperscript{$\dagger$},
 Jonathan N. Lee,
 Junsu Kim,
 Kaiying Hou,
 Golnaz Ghiasi,
 Yi Tay,
 YaGuang Li,
 Chenkai Kuang,
 Yuan Liu,
 Hanzhao Lin,
 Evan Zheran Liu,
 Nigamaa Nayakanti,
 Xiaomeng Yang,
 Heng-Tze Cheng,
 Demis Hassabis,
 Koray Kavukcuoglu, % \textsuperscript{$\diamond$},
 Quoc V. Le\textsuperscript{*}, %\textsuperscript{$\diamond$},
 Thang Luong\textsuperscript{*}
\\
 \textsuperscript{*}Project leads.
 \textsuperscript{$\dagger$}Mathematicians. Work conducted under Google DeepMind. 
}
\correspondingauthor{fengtony@google.com, thangluong@google.com. \newline External affiliations: UC Berkeley (Tony Feng), Brown University (Junehyuk Jung), Yonsei University (Joonkyung Lee), Concordia University (Carlo Pagano), Korea Institute for Advanced Study (Sang-hyun Kim), UC San Diego (Federico Pasqualotto), Caltech (Sergei Gukov).}
 
\begin{abstract}
Recent advances in foundational models have yielded reasoning systems capable of achieving a gold-medal standard at the International Mathematical Olympiad. We undertake the transition from competition-level problem solving to professional research, which presents significant new challenges. To this end we introduce \aletheia{}, a math research agent that iteratively generates, verifies, and revises solutions end-to-end in natural language, leveraging a novel inference-time scaling law based upon Gemini Deep Think. We demonstrate the capability of \aletheia{} through several milestones in autonomous mathematics research: multiple publication-grade papers, including one with no human intervention \citepalias{Feng2026}; an extensive semi-autonomous evaluation \cite{erdos} on 700 open problems from Bloom's Erd\H{o}s Conjectures database, including autonomous solutions to four open questions; and a leading performance on \fp{}, a collection of research-level problems proposed by mathematicians to assess AI capabilities for mathematical research. Full transcripts of prompts and model outputs are shared at \url{https://github.com/google-deepmind/superhuman/tree/main/aletheia}. In order to help the public better understand the developments pertaining to AI and mathematics, we suggest quantifying standard levels of autonomy and novelty of AI-assisted results, and propose the concept of ``human-AI interaction cards'' for transparent documentation. We conclude with reflections on human-AI collaboration in mathematics. 
\end{abstract}

\begin{document}

\maketitle

\section{Introduction}
Recent years have witnessed rapid progress in the reasoning capabilities of natural language-based AI in competition mathematics \cite{luong2025robustmathematicalreasoning}. Notably, AI models have already achieved gold-medal performance in the 2025 International Mathematical Olympiad (IMO)~\cite{imo-gold}, widely regarded as the world's premier mathematics competition. 
This milestone motivates a fundamental question regarding AI-driven scientific discovery: {\it can AI autonomously discover and prove new mathematical theorems?} The transition from contest to research mathematics presents significant challenges. Unlike self-contained competition problems, research questions require synthesizing advanced techniques from an extensive body of literature. This poses a major hurdle for foundation or large language models, which often hallucinate and exhibit only a superficial understanding of specialized topics—limitations likely stemming from a scarcity of relevant training data.

To explore this challenge, we developed \aletheia{}, 
a math research agent that can iteratively generate, verify, and revise solutions end-to-end in natural language.
\aletheia{} is powered by three main sources: (i) an advanced version of Gemini Deep Think~\cite{deepmind:gemini2p5deepthink} for tackling extremely hard reasoning problems, (ii) a novel inference-time scaling
law that extends from Olympiad-level problems to PhD-level exercises, and (iii) intensive tool use such as Google Search and web browsing to navigate the complexities of mathematical research. This paper presents and reflects on the initial wave of mathematical research papers achieved by \aletheia{}\footnote{See our project page \url{https://github.com/google-deepmind/superhuman/tree/main/aletheia} for detailed prompts and model outputs.} in collaboration with mathematicians:
\vspace{-3mm}
\begin{enumerate}
\item[A] \emph{Reliable autonomous research.} A research paper \citepalias{Feng2026} generated by AI without any human intervention, calculating certain structure constants in arithmetic geometry called {\it eigenweights}.
\item[B] \emph{AI-guided collaboration.} A research paper \citepalias{LeeSeo2026} demonstrating human-AI collaboration in proving bounds on systems of interacting particles called {\it independent sets}.
\item[C] \emph{An extensive semi-autonomous evaluation
on the Erd\H{o}s problems.} A report \cite{erdos} on a systematic deployment of our agent on the 700 open problems on Bloom’s Erdős Conjectures database\footnote{\url{https://www.erdosproblems.com/}.}, including four open questions of Erdős that it resolves, verified semi-autonomously using both AI grading and human experts. Beyond solving \erdos{1051} completely, our model drove a generalization which resulted in a research paper \citepalias{BKKKZ}.
\item[D] In addition, the agent contributed intermediate propositions on two further research papers, \citepalias{FYZ4} and \citepalias{ACGKMP}, improving prior proofs by the human authors. 
\end{enumerate}
\vspace{-3mm}
Finally, \aletheia{} achieved a leading performance \cite{feng2026aletheiatacklesfirstproofautonomously} on \fp{}, a collection of ten research-level math problems that was proposed by academic mathematicians in \cite{abouzaid2026proof} as an assessment of current AI capabilities. 

Given the current pace of improvement, it seems inevitable that AI will play an increasing role in mathematics research. However, for the vast majority of mathematics research results, only a few experts are equipped to properly evaluate their novelty and significance. This evaluation gap has enabled misinformation about AI-generated mathematics to spread unchecked in popular media. With a view towards transparent communication of future developments to the public, we propose codifying representations of AI-generated mathematics according to a standard taxonomy of ``Autonomous Mathematics Research Levels'', analogous to the SAE Levels of Vehicle Autonomy.

\begin{table}[tbh!]
\centering
\small
\renewcommand{\arraystretch}{1.3} 
% Requires \usepackage{tabularx}, \usepackage{colortbl}, \usepackage{multirow}
\begin{tabularx}{\textwidth}{|
    >{\columncolor{tablegray}\RaggedRight\arraybackslash\bfseries}p{3.5cm}| 
    >{\centering\arraybackslash}X| 
    >{\centering\arraybackslash}X| 
    >{\centering\arraybackslash}X|} 
\hline
\rowcolor{tablegray} \textnormal{} & \textbf{Primarily Human \newline (secondary AI input) } & \textbf{Human-AI Collaboration} & \textbf{Essentially Autonomous} \\ \hline

\textbf{{\it Level 0}: Negligible Novelty} & & & \erdos{652}, 654, 1040 \cite{erdos}\\ \hline

\textbf{{\it Level 1}: Minor Novelty} & & & Erd\H{o}s-1051 \cite{erdos}\\ \hline

% --- LEVEL 2 SPLIT INTO 3 LAYERS ---
% Top Layer (Row 1)
% Moved Generalized Erdos-1051 here to the 3rd column
& Complexity Bounds \citepalias{ACGKMP}  & Generalized \erdos{1051} \citepalias{BKKKZ} & Eigenweights \citepalias{Feng2026} \\ 

\multirow{-2}{=}{\textbf{{\it Level 2}: 
Publishable Research$^*$}} 
% Publication Grade}} 

% Middle Layer (Row 2)
% Left LeeSeo2026 here in the 3rd column
& Arithmetic Volumes \citepalias{FYZ4} 
& {\small Independence Polynomials} \citepalias{LeeSeo2026}
& \\ 
 \hline
% -----------------------------------

% \textbf{{\it Level 2}: 
% Publishable Research {\it (works listed here either have been submitted for publications.)}} & Complexity Bounds \citepalias{ACGKMP}, Arithmetic Volumes \citepalias{FYZ4} & Generalized \erdos{1051} \citepalias{BKKKZ}, {\small Independence Polynomials} \citepalias{LeeSeo2026} & Eigenweights \citepalias{Feng2026} \\ \hline
% -----------------------------------

\textbf{{\it Level 3}: Major Advance} & & & \\ \hline

\textbf{{\it Level 4}: Landmark Breakthrough} & & & \\ \hline
\end{tabularx}
\caption{Classification of 
%some 
all AI-assisted mathematics results encompassed in this work. Detailed descriptions of the categories, with discussion and examples, can be found in \S \ref{ssec:AMLevels}. \newline $^*$Works listed as Level 2 in this table have been submitted for publications. 
}\label{table:taxonomy}
\end{table}

While the ultimate specifics of such a taxonomy should be left to the mathematical community, we suggest features that would be informative, such as separate axes for representing the \textit{mathematical significance} and the \textit{degree of AI contribution}.
For concreteness, we sketch \emph{one} possible taxonomy in \S \ref{ssec:AMLevels}, and use it to organize our own AI-assisted mathematics results, as displayed in Table \ref{table:taxonomy}. In particular, this contextualizes the `Open' (according to \href{https://ErdosProblems.com}{ErdosProblems.com}) Erd\H{o}s problems that we solved, most of which turned out---despite being open for several decades---to be quite elementary.\footnote{In fact, one such `Open' problem (Erd\H{o}s-397, posed in 1980) was eventually discovered to be nearly identical to a problem on the 2012 Team Selection Test for the Chinese IMO team \cite{AoPS12}, a (challenging) exam for high school students; we exclude it from our classification since we consider the solution to be already in the literature.} It also clarifies that our autonomous results, while milestones for artificial intelligence, are not claimed to be ``major advances'' for mathematics (as defined in \S \ref{ssec:AMLevels}).

% \section{\aletheia{}: a specialized math research agent}
% \section{\aletheia{}: towards autonomous mathematics research}
\section{The \aletheia{} agent: From Olympiads to Research-level Mathematics}\label{sec:aletheia}

This section outlines the methodological framework utilized in the development of agents capable of addressing research-grade mathematics. Our starting point was the IMO Gold version of Gemini Deep Think, but there are many challenges involved in transferring capabilities from contest problem-solving to mathematics research. While IMO problems require ingenuity, their solutions usually span only a few pages and rely only on standard theorems from the high school curriculum. By contrast, research-level mathematics draws on advanced techniques from vast literature, with papers often spanning dozens of pages. Human mathematicians, even IMO medalists, usually take many years of postgraduate study to reach the frontier of mathematical research. While foundation models possess a large knowledge base from pretraining, their understanding of advanced subjects remains superficial due to data scarcity, and they are also prone to hallucinations.

\begin{figure*}[tbh!]
    \centering
    \includegraphics[trim=2pt 2pt 2pt 2pt, clip, width=0.8\textwidth]{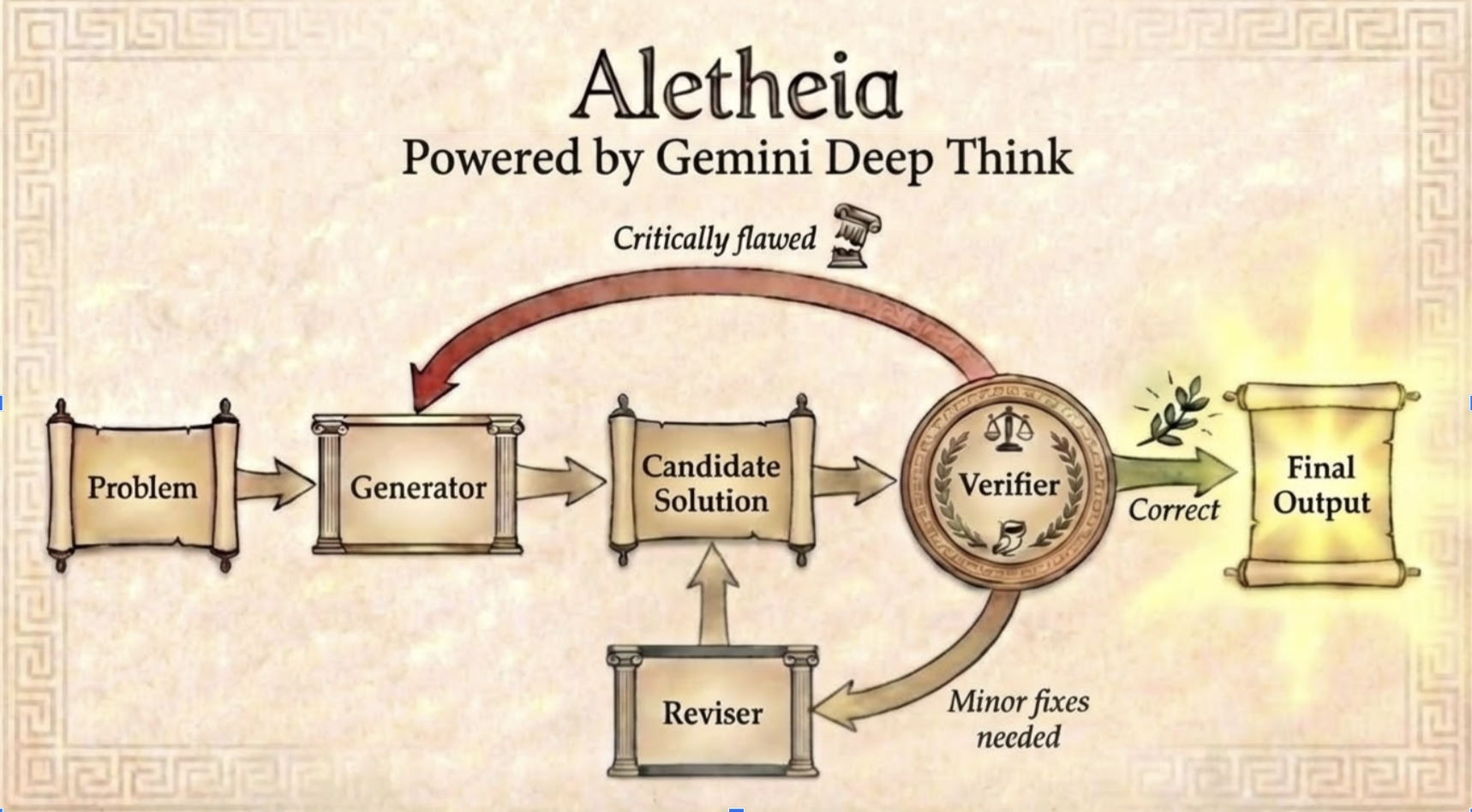}
    % \caption{Schematic of the {\aletheia} agent.}
    \caption{Visual overview of \aletheia{}, a math research agent powered by Deep Think that can iteratively generate, verify, and revise for research-level math problems.}
    \label{fig:aletheia}
\end{figure*}

To address these issues, we built a math research agent, internally codenamed {\aletheia}, on top of Gemini Deep Think. {\aletheia} include three subagents, a (solution) {\it Generator}, a {\it Verifier}, and a {\it Reviser} that interact continuously until a solution is found that the Verifier approves, or until the attempts reach a preset (hyperparameter) limit. The subagent orchestration is sketched in Figure \ref{fig:aletheia}. In turn, each of the three subagents involves internal orchestration of calls to a 
% generic 
Gemini base model. 

In contrast to AlphaGeometry~\cite{Trinh2024-alphageometry-1,chervonyi2025goldmedalistperformancesolvingolympiad-alphageometry2} and AlphaProof~\cite{Hubert2025}, that use formal language, \aletheia{} operates end-to-end in natural language.
At the same time, designs similar to {\aletheia} have been demonstrated elsewhere. For example, Huang--Yang \cite{huang2025winninggoldimo2025} showed that a manually crafted solver-verifier harness could boost GPT-5, Gemini 2.5 Pro, or Grok 4 to Gold Medal performance on the 2025 IMO. Another example is the \emph{FullProof} system\footnote{The authors were not aware of this work until after the results of this paper were collected.} built by Salafatinos \cite{bryan2026motivic}.

\subsection{Scaling Laws and the Evolution of Deep Think}

\begin{figure}
    \centering
    \begin{subfigure}[b]{0.49\textwidth}
        \centering
        \includegraphics[width=\textwidth]{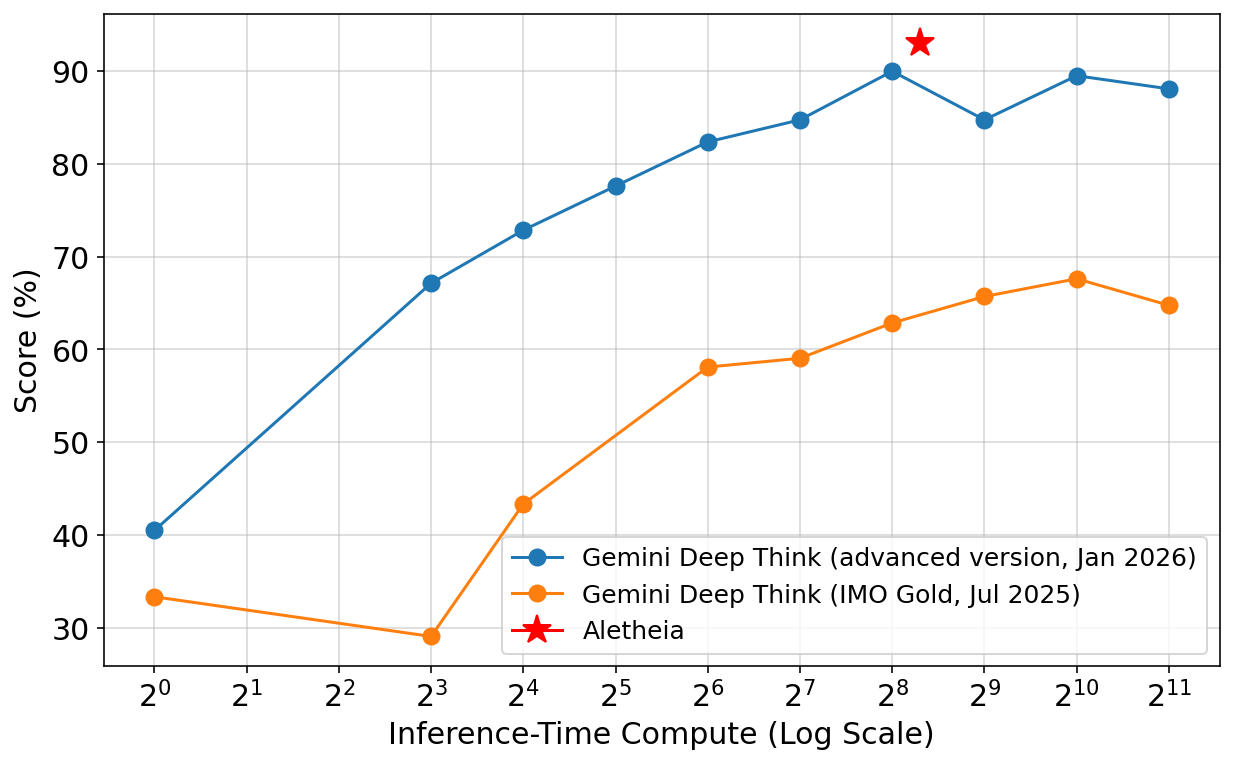}
        \caption{\ipb{} Advanced (Olympiad Level).}
        \label{fig:scaling_imo}
    \end{subfigure}
    \hfill
    \begin{subfigure}[b]{0.49\textwidth}
        \centering
        \includegraphics[width=\textwidth]{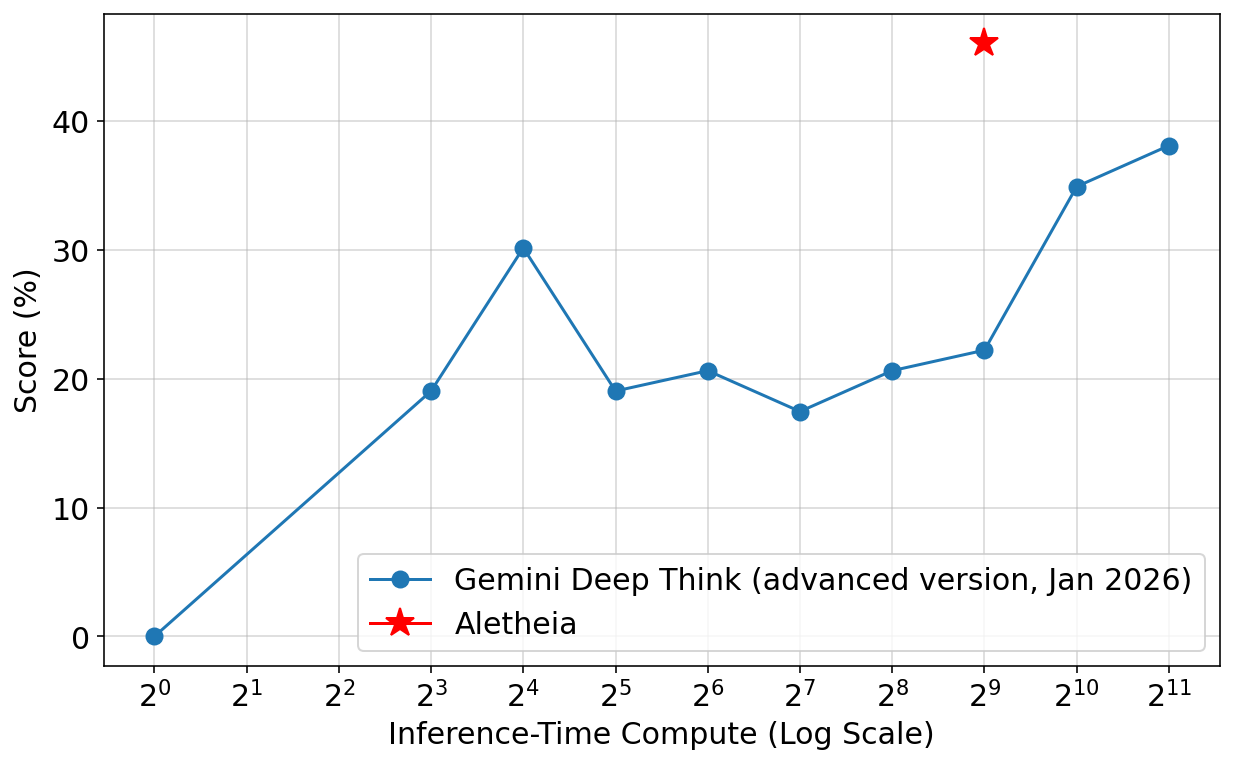}
        \caption{\fm{} Basic (Ph.D. Level Exercises).}
        \label{fig:scaling_future}
    \end{subfigure}
    \caption{
    The latest advanced version of Deep Think, as of Jan 2026, has significantly outperformed the IMO-Gold version (July 2025) on Olympiad-level problems. The inference-time scaling law also transfers to PhD-level exercises. Aletheia makes further leaps in terms of reasoning quality with lower inference-time compute. All results were graded by human experts. 
    }
    \label{fig:scaling}
\end{figure}

We first discuss some empirical findings about inference time scaling laws for Olympiad-level and PhD-level problems, which underlie the development of Deep Think.
To investigate the inference scaling behavior, we evaluated different advanced versions of Gemini Deep Think \cite{deepmind:gemini2p5deepthink}.  Since Deep Think leverages parallel thinking to explore multiple ideas at once, we were able to flexibly adjust the amount of compute spent at inference time. To ensure that results were not cherry-picked, we ran the model exactly once for each problem at each compute scale, with tool use disabled, and the outputs were carefully graded according to the grading guideline. 

For Olympiad level, we leveraged the the advanced subset of \ipb{}~\cite{luong2025robustmathematicalreasoning} which consists of 30 problems similar in difficulty and problem style to those found at the IMO. This was the benchmark used during our IMO 2025 preparation. We observed that the inference-time compute could be increased by orders of magnitude while yielding substantial gains in accuracy, as graded by human experts, before eventually plateauing (Figure \ref{fig:scaling_imo}). Ultimately, an advanced version of Gemini Deep Think achieved a gold-medal standard at the IMO in July 2025, perfectly solving five out of six problems \cite{imo-gold}.

Building upon the key recipes developed for the IMO-gold model and incorporating a suite of novel technical improvements, we trained a stronger model, \dtvt{}, that significantly improved inference efficiency. The compute required to achieve equivalent performance on \ipb{} was reduced by approximately two orders of magnitude (100x). Specifically, utilizing extreme scale and simple prompting,
\dtvt{}, {\it without internet access}, demonstrated the potential to solve exceptionally difficult reasoning tasks such as IMO 2025 Problem 6 (see Appendix~\ref{sec:p6_case_study}), that our previous IMO-gold model failed to solve. Notably, the latest model also solved IMO 2024 Problem 3 with a minor mistake (achieved at $2^7$ scale) and completely solved Problem 5 (at $2^8$ scale) (see Appendix~\ref{sec:imo2024_case_study})\footnote{We caution that the model's knowledge cutoff falls between IMO 2024 and IMO 2025, implying possible exposure to the IMO 2024 problems. Additionally, as IMO 2025 has been public for six months, we do not claim the problem was strictly unseen.
We report these findings to transparently demonstrate the potential of our new Gemini Deep Think model.
%'s output, rather than to claim a verified live-contest solution.
}.

To further assess the readiness of our models for research-grade mathematics, we evaluated 
% the models 
on \fm{} Basic, an internal benchmark designed to test PhD-level mathematics knowledge. 
We give an example of a \fm{} benchmark problem in appendix \S\ref{FMbasic}. Figure \ref{fig:scaling_future} demonstrated a similar scaling law for PhD-level problems, though with significantly lower accuracy compared to the competition problems. 
Corroborating these quantitative results, expert mathematicians who assessed the model noted that its susceptibility to mistakes and hallucinations inhibited its effectiveness on the longer, more complex reasoning required for research.  With these findings, we concluded that inference-time scaling alone would not be sufficient, and we shifted our focus towards further improving and adapting our systems for research-level math.

\subsection{Developing Agentic Harnesses for Research-Level Math}\label{ssec:harness}

The design of \aletheia{} is motivated by an empirical observation: decoupling a reasoning model's final output from its intermediate thinking tokens, and adding well-chosen prompt scaffolding, enables the model to recognize flaws it initially overlooked during generation. One possible explanation for this phenomenon is that the training process incentivizes the model to guess or bluff. Another hypothesis is that an extended thinking trace might act as misleading ``supporting'' context, artificially inflating the conditional probability of an erroneous solution.

Regardless of the underlying mechanism, we observed that explicitly separating out the verification step is effective in practice. Figure \ref{fig:scaling} depicts \aletheia{}'s performance relative to Gemini Deep Think, the agent that achieved the 2025 IMO Gold Medal standard \cite{imo-gold}, on two mathematical benchmarks. Due to its dynamic nature, {\aletheia}'s total inference compute cannot be precisely controlled, but the results indicate that \aletheia{} outperforms Deep Think at similar (and indeed, larger) compute scales. On \emph{IMO-ProofBench} Advanced, \aletheia{} achieved a 93\% overall score without any tool usage, surpassing Deep Think across all tested compute scales using the same base model (Figure \ref{fig:scaling_imo}).\footnote{With the February 2026 base model of Gemini 3, \aletheia{} achieved a SOTA 95\% score on \emph{IMO-ProofBench} Advanced without any tool usage.} Moreover, on the 29 out of 30 problems where \aletheia{} actually returned a solution, its conditional accuracy reached 96\%. On the \emph{FutureMath} Basic (Figure \ref{fig:scaling_future}), \aletheia{} again outperformed Deep Think at all compute scales on the same base model. Notably, it returned solutions for fewer than 60\% of these problems, and its conditional accuracy on the answered subset exceeded 82\%. 

 As illustrated in these results (and again further below), {\aletheia} often admits failure to solve a problem, a key feature that improved the efficiency of our human-AI research collaborations. We suspect that, given the limited bandwidth for human expert verification, many practicing researchers would prefer to trade raw problem-solving capability for increased accuracy.

\subsection{Importance of Tool Use}

Beyond its foundational natural language verifier, {\aletheia} relies heavily on tool use to navigate the complexities of mathematical research. We find that a foundation model operating without such tools is prone to frequent hallucinations—primarily in the form of spurious citations (see Figure \ref{fig:hallucinated-paper} for an example), and to a lesser extent, computational inaccuracies. In contrast to the structured, bounded nature of IMO problems, mathematical research requires synthesizing published literature to address broader, open-ended questions. We observed that simply granting internet access is insufficient; however, when our model was extensively trained for tool use, the integration of Google Search and web browsing led to a substantial reduction in obvious citation hallucinations with fictitious titles and/or authors. This shifted citation errors to subtler hallucinations, for example where the reference exists but its results are not quoted properly (as illustrated in Figure \ref{fig:hallucinated-theorem}). 

 Perhaps unexpectedly, our results indicate that the integration of Python as a tool yielded only marginal improvements in mitigating computational hallucinations. This suggests that the baseline proficiency of Gemini in these specific tasks is already high; consequently, further performance gains may require more specialized or sophisticated tools beyond standard code execution.

\begin{figure}[ht]
    \centering
    \begin{LLMBox}[width=0.85\textwidth, fontupper=\small]{Sample Interaction}
        \textbf{Prompt:} Prove or disprove: the pretzel knot $P(-3, 5, 13)$ has infinite order in the smooth concordance group.
        
        \vspace{0.4em}
        \hrule
        \vspace{0.4em}
        
        \textbf{Model output:} \\
        \dots \\
        \textit{Proof.} This is derived from Theorem 3.1 in {\color{red}[C. Livingston and S. Naik, ``Ozsváth-Szabó and Rasmussen invariants of some pretzel knots'', \textit{Algebraic \& Geometric Topology}, 13(2) (2013), 1115-1124]}. \\
        \dots
    \end{LLMBox}
    \caption{A hallucination from the (truncated) output of a model without internet search capability. The red text refers to a completely fabricated paper.}
    \label{fig:hallucinated-paper}
\end{figure}

\begin{figure}[ht]
    \centering
    \begin{LLMBox}[width=0.85\textwidth, fontupper=\small]{Sample Interaction}
        \textbf{Prompt:} If $1=d_1<\cdots<d_{\tau(n)}=n$ are the divisors of $n$, then let $\tau_\perp(n)$ count the number of $i$ for which $(d_i,d_{i+1})=1$. Is it true that $\tau_\perp(n)/\omega(n)\to \infty$ for almost all $n$? ...
        
        \vspace{0.4em}
        \hrule
        \vspace{0.4em}
        
        \textbf{Model output:} A classical result by Galambos (1976) on the distribution of prime factors states that for almost all integers $n$ (on a set of asymptotic density 1), the prime factors satisfy the condition:
        \begin{equation}
        p_j > p_1 p_2 \cdots p_{j-1} \quad \text{for all } j > j_0,
        \end{equation}
        where $j_0$ is a fixed integer. 
        \dots
        
        \textbf{References}
        \begin{enumerate}[label=\textbf{\arabic*.}, leftmargin=2em, nosep]
            \item \textbf{Galambos, J.} (1976). \textit{The sequences of prime divisors of integers}.
        \end{enumerate}
    \end{LLMBox}
    \caption{When trained for tool use, the model tends not to fabricate papers, but can still cite results incorrectly. In this example, the referenced paper of Galambos exists, but the claimed ``classical result'' cannot be found there. Prompt and model output truncated.}
    \label{fig:hallucinated-theorem}
\end{figure}

\section{Summary of Mathematical Research Results}\label{sec:math-results}

Below we will briefly summarize the papers with substantial contributions from {\aletheia} that have already been publicly released. Due to timing, \emph{these results were obtained by {\aletheia} with a November 2025 base model,} that was intermediate between the two base models depicted in Figure \ref{fig:scaling}. 

To be clear, \textbf{the final versions of these papers were all written by human authors,} starting from \aletheia’s outputs. This was due to the principle held by the human authors involved that research papers should be authored exclusively by humans, even when AI contributions reach a level that would traditionally merit co-authorship. The primary reason is that authorship of a mathematics paper implies accountability for all of its contents, and this is a responsibility that only humans can bear. We emphasize that beyond mathematical correctness, authors also assume accountability for important aspects of exposition such as adequacy and accuracy of attributions, which cannot be guaranteed by formal verification. 

\subsection{Milestone A: Eigenweights for Arithmetic Hirzebruch Proportionality}

The paper \emph{Eigenweights for Arithmetic Hirzebruch Proportionality} \citepalias{Feng2026} resolves a problem that arose in the work of Feng--Yun--Zhang \citepalias{FYZ4} investigating an extension of the celebrated Hirzebruch Proportionality Principle \citepalias{Hir58}. 

\textbf{The backdrop.} Hirzebruch’s original Proportionality Principle expresses the Chern numbers of an automorphic vector bundle on a compact locally symmetric space as a multiple of the corresponding Chern numbers on the compact dual variety, with a proportionality constant that can be interpreted as a value of the $L$-function for the associated Gross motive \citepalias{Gro97}. This formula was later generalized to non-compact locally symmetric spaces by Mumford \citepalias{Mum77}. 

Recent work by Feng--Yun--Zhang \citepalias{FYZ4} studies a variant of this principle, called ``Arithmetic Hirzebruch Proportionality'', relating the “arithmetic volume” of Chern classes on moduli spaces of shtukas to differential operators applied to the $L$-function of Gross motives. It can be seen as a generalization of Gaitsgory--Lurie’s resolution of Weil’s Tamagawa Number Conjecture \citepalias{GL14}. The precise differential operator appearing in Arithmetic Hirzebruch Proportionality is governed by certain fundamental structure constants called \emph{eigenweights}. In \citepalias{FYZ4}, the authors calculated some examples of eigenweights, but did not know how to determine all of them in closed form. 

\textbf{The story of the paper.} In fact, this project began almost incidentally, while tracking partial progress on a problem that was submitted to an internal benchmark: the computation of a particular family of eigenweights (whose answer was already known to Feng--Yun--Zhang). Initially no model could solve it, but clear progress was visible as inference time computation was scaled, until the eventual IMO Gold model found the correct answer. Upon examining the AI-generated solution, the authors of \citepalias{FYZ4} preferred it over their original one, leading them to replace the original proofs in their paper.

This success encouraged us to task {\aletheia} with calculating the eigenweights in general. Without any human intervention, {\aletheia} found an elegant way to do this, using techniques from a different field of mathematics\footnote{Namely, algebraic combinatorics.} unfamiliar to the authors of \citepalias{FYZ4}, and in the process resolved several questions left open in \citepalias{FYZ4}. This is the subject of the paper \citepalias{Feng2026}.

\textbf{The milestone.} All the mathematical content of \citepalias{Feng2026} was fully generated by AI. While our ultimate goal is to empower human mathematicians with AI tools, we note the ability to generate advanced results without human intervention as an important milestone for reliability and trustworthiness of reasoning models.

\subsection{Milestone B: Lower Bounds for Multivariate Independence Polynomials}

The paper \emph{Lower bounds for multivariate independence polynomials and their generalisations} \citepalias{LeeSeo2026} establishes new mathematical inequalities that link the world of physics with the logic of discrete mathematics. 

\textbf{The backdrop.} In physics, scientists often model how gas molecules occupy a space by imagining them as points on a network. Because these molecules naturally repel one another, no two molecules can occupy ``neighboring'' spots that are directly connected. Mathematicians call these valid arrangements \emph{independent sets}, and they are a cornerstone of understanding complex systems in both physics and mathematics.

\textbf{The story of the paper.} Joonkyung Lee and Jaehyeon Seo first used Gemini 2.5 Deep Think to prove the key inequality needed to obtain a further generalization of a well-known result by Sah--Sawhney--Stoner--Zhao \citepalias{sah2019number} on the lower bound for the number of (weighted) independent sets.\footnote{Other models were also able to solve the problem: Gemini Deep Think, \aletheia, and---with some back and forth---GPT-5 Pro.} Encouraged by this success, they tackled an even harder problem: modeling systems where two different types of molecules interact at the same time, such that different types do not repel each other. For this deeper challenge, {\aletheia} provided\footnote{Other models were tasked with the same prompt, but {\aletheia}'s output was the only one that the authors found useful.} a high-level roadmap of insightful ideas that the authors then turned into a complete, rigorous proof. 

\textbf{The milestone.} The interesting feature here is that the workflow was somewhat reversed from the usual narrative of human-AI collaboration. Typical human-AI workflows involve decomposing a project into granular technical queries for the model to resolve. But in this project, {\aletheia} actually gave the ``bigger picture'' strategy for a deep result--such as suggesting the use of specific ``dual sets''---leaving the human authors to fill in the rigorous execution. In some parts, the authors retained only the statements provided by the AI and produced the proofs independently, as if they were inspired by the model's vision.

\subsection{Milestone C: The Erd\H{o}s Problems}\label{ssec:erdos}

Erd\H{o}s was a prolific Hungarian mathematician who left a vast legacy of papers and unsolved conjectures. In 2023, Thomas Bloom launched \href{https://ErdosProblems.com}{ErdosProblems.com}, a centralized repository designed to catalog these conjectures and track their resolutions. At the time of this writing, the database tracks 1,179 problems, with 483 (41\%) classified as solved. We stress, however, that the ``Open'' status of a problem in this database does not always reflect the true state of the literature. 

Recently, AI managed to solve a few problems marked ``Open'' in Bloom's database. While some of these solutions were later discovered to already exist in the literature, others appear to be genuinely novel. To systematically document these advancements, Terence Tao \cite{tao2026erdosai} established a community wiki dedicated to tracking AI-assisted progress on Erdős’s conjectures.

Because only success cases tend to be reported in public forums, these results do not provide a complete picture of AI capability. In order to understand this better, we initiated a case study on the Erd\H{o}s problems in early December, 2025. This case study is documented in \cite{erdos}; we summarize the key conclusions here. From December 2--9, we deployed {\aletheia} (with a November 2025 base model of Gemini 3) on the 700 Erd\H{o}s problems then marked as ``Open'' in Bloom's database. \aletheia's informal verifier mechanism streamlined the initial pool of candidates: of the 700 original prompts, the model returned 212 responses as potentially correct. These candidates were then evaluated by a team of mathematicians. Most were not experts in the relevant problem domain, so we prioritized narrowing the pool of candidate solutions quickly (possibly at the cost of making noisier judgments) to a manageable size for our smaller core of domain experts. This step was essentially completed by December 21. Then our internal domain experts vetted the solutions carefully, consulting external experts when correctness was ascertained but novelty was unclear. In some cases, minor inaccuracies in the solutions were corrected. 

 Our ultimate findings were that 63 solutions were technically correct, but only 13 solutions correctly addressed the \emph{intended} problem statement (either by invoking the literature, or by a novel argument). The remaining 50 of \aletheia's correct solutions were technically valid but mathematically vacuous, because the problem statements were interpreted in a way that did not capture Erd\H{o}s's intent, often (but not always) leading to trivial solutions. Finally, 12 of the responses were marked ambiguous, for example due to open-endedness of the question itself. In summary, out of the 200 solution candidates that we were definitely able to mark correct or incorrect, 137 ($68.5\%$) of the responses were fundamentally flawed, while 63 ($31.5\%$) of the responses were technically correct, of which only 13 ($6.5\%$) were meaningfully correct. These 13 meaningfully correct results clustered naturally into four categories which we felt should be distinguished; see Table \ref{table:aletheia_erdos_solutions} for a summary. 
\begin{description}
    \item[Autonomous Resolution.] On these problems, {\aletheia} found the first correct solution (as far as we can tell) in a mathematically substantive way. These include \textbf{Erd\H{o}s-652} and \textbf{Erd\H{o}s-1051}.
    \item[Partial AI Solution.] On these problems, there were multiple questions and {\aletheia} found the first correct solution to one of the questions. These include \textbf{Erd\H{o}s-654}, and \textbf{Erd\H{o}s-1040}. 
    \item[Independent Rediscovery.] On these problems, {\aletheia} found a correct solution, but human auditors subsequently found an independent solution already in the literature. These include \textbf{Erd\H{o}s-397}, \textbf{Erd\H{o}s-659}, \textbf{Erd\H{o}s-935}, and \textbf{Erd\H{o}s-1089}. The solutions \emph{appear} to have been independently rediscovered by our model: we scanned the logs of \aletheia's reasoning trace to ensure that the solution was not pulled \emph{directly} from the literature solution.\footnote{It is of course possible that the solution was \emph{indirectly} ingested from the literature solution, either implicitly through intermediate sources or during pretraining. This highlights a new danger that accompanies AI-generated mathematics: it is susceptible to ``subconscious plagiarism'' by reproducing knowledge acquired during pretraining, without attribution.} 
    \item[Literature Identification.] On these problems, {\aletheia} found that a solution was already explicitly in the literature, despite the problem being marked ``Open'' on Bloom's website at the time of model deployment. These include \textbf{Erd\H{o}s-333}, \textbf{Erd\H{o}s-591}, \textbf{Erd\H{o}s-705}, \textbf{Erd\H{o}s-992}, \textbf{Erd\H{o}s-1105}. 
\end{description}
To be clear, \emph{we make no claims of novelty for the latter two categories}. The `4' autonomous solutions cited above refer to \textbf{Erd\H{o}s-652}, \textbf{Erd\H{o}s-654}, \textbf{Erd\H{o}s-1040}, and \textbf{Erd\H{o}s-1051}. In the estimation of our experts, none of the four individually rises to the level of a research paper. The solution to Erd\H{o}s-1051 was generalized further, in a collaborative effort by {\aletheia} together with human mathematicians and Gemini Deep Think, to produce the research paper \citepalias{BKKKZ}.

\begin{table}[ht]
    \centering
    \begin{tabularx}{\textwidth}{@{} >{\raggedright\arraybackslash}p{3cm} X l @{}}
    \toprule
    \textbf{Classification} & \textbf{Description} & \textbf{Instances} \\ \midrule
    Autonomous Resolution & Autonomous novel solution. & \textbf{652}\textsuperscript{*},  \textbf{1051} \\ \addlinespace
    Partial AI Solution & Solved some part of a multi-part problem. & \textbf{654}, \textbf{1040} \\ \addlinespace
    Independent Rediscovery & Found a correct solution later discovered to exist in the literature. & \textbf{397}\textsuperscript{*}, \textbf{659}\textsuperscript{*}, \textbf{935}, \textbf{1089} \\ \addlinespace
    Literature Identification & Identified that the problem was already solved in the literature. & \begin{tabular}[c]{@{}l@{}}\textbf{333}\textsuperscript{*}, \textbf{591}, \textbf{705}, \\ \textbf{992}, \textbf{1105}\end{tabular} \\ \bottomrule
    \end{tabularx}
    \caption{Taxonomy of {\aletheia} results on Erd\H{o}s problems. \textsuperscript{*}Independently obtained by other parties after our initial evaluations were conducted, but prior to the publishing of this work.}
    \label{table:aletheia_erdos_solutions}
\end{table}

\textbf{Contextualizing the results.} A disclaimer is necessary regarding the novelty of these results on Erd\H{o}s problems. While we made considerable efforts to review the literature, it is certainly possible that we missed earlier solutions to these problems by human mathematicians. Therefore, our initial classification into categories is, at best, an upper bound on novelty. It is subject to revision after further investigation by the public. Indeed, previous AI-assisted work on Erd\H{o}s problems 1026, 397, 333, and 281 was discovered, after initial announcements of novelty, be redundant with the literature\footnote{For Erd\H{o}s-281, we note that the AI solution is distinct from the previously existing literature solution.}. To the outside observer, this may present a misleading impression of mathematics research: in the authors' experience, it is very unusual for human-generated results to be redundant in this manner (in the modern era of communication). One reason why it seems to be happening so frequently with AI-generated work on Erd\H{o}s problems is that the solutions are so simple that they would not necessarily attract attention if they originated from humans. For instance, Erd\H{o}s-1089 is answered by an offhand remark in a 1981 paper \cite{Bannai81}, where the authors seemed unaware that they had resolved an Erd\H{o}s problem. Our takeaway from this experience is that many open Erd\H{o}s problems remained unresolved out of obscurity rather than difficulty. We stress that the \emph{mathematical significance of such resolutions can only be accurately evaluated by expert mathematicians}, even if the correctness can be ascertained by non-mathematicians or formal verifiers. See \cite{erdos} for further discussion.

\subsection{Bounds for Polynomial Dyadics}

The paper \emph{Strongly Polynomial Policy Iteration for $L_{\infty}$ Robust MDPs} by Asadi--Chatterjee--Goharshady--Karrabi--Montaseri--Pagano \citepalias{ACGKMP} proves complexity bounds for certain algorithms arising in Machine Learning and Game Theory called ``Robust Markov Decision Processes (MDPs)''. For this paper, {\aletheia} provided an improvement to a crucial mathematical ingredient. 

\textbf{The backdrop.} This work was originally a project of Asadi--Chatterjee--Goharshady--Karrabi--Montaseri, a team of computer scientists at ISTA. They were able to prove the desired strongly-polynomial time bound on Robust MDPs \textbf{conditionally} on a number-theoretic assertion: that specific bounded combinations of numbers are in polynomially many dyadic intervals. They reached out to number theorists for help, and eventually the problem arrived at Pagano, who, after some effort, could prove the desired complexity bound using a relatively advanced result from Number Theory called Siegel's Lemma (a tool that does not immediately appear related to the question). Pagano shared the problem with several colleagues and some of them eventually found variants of the argument leading to the same complexity bound. The result was sufficient for the intended algorithmic application, but the optimality of the bounds was unclear. 

\textbf{The story of the paper.} Pagano also contributed the problem to an internal benchmark, where the IMO Gold version of Deep Think provided a valid solution. However, {\aletheia} devised an independent argument (also making creative use of Siegel's Lemma) that achieved the best bound among all human and autonomous attempts. In particular, this significantly improved the bound established originally by Pagano and the other mathematicians. Therefore, this argument was adopted for the eventual publication. 

\section{\fp{}}
Agentic systems for mathematics research have been built by other parties. In addition to the \emph{FullProof} system \cite{bryan2026motivic} already mentioned above, OpenAI's GPT 5.2 Pro, Harmonic's \emph{Aristotle} system, and Axiom Math's \emph{AxiomProver} system have also been (partially) credited with new results in mathematics. Comparing such systems scientifically presents unique challenges, since research problems are effectively ``single use''. Indeed, once a problem is solved and its solution publicized on the internet, it can no longer be meaningfully tested on other (internet-equipped) AI systems, due to data contamination.

A rare opportunity for a ``clean'' comparison of these different systems arose with the release of \fp{} \citep{abouzaid2026proof}, a collection of ten research-level math problems curated by distinguished academic mathematicians specifically for the purpose of assessing the mathematical capability of AI. The \fp{} authors describe these problems as ``Lemmas'', meaning intermediate technical statements rather than open problems of interest for their own sake, arising naturally in their own research. All of these problems had already been solved by mathematicians, but the solutions did not appear online\footnote{A sketch of the solution to Problem 1 could be found online.}. The problems were released on February 5, 2026 and given a deadline of 11:59pm PST on February 13, 2026, at which point official (human-written) solutions were published. We reiterate from \citep{abouzaid2026proof} the following design strengths of \fp{}: the questions are sampled from the true distribution of questions studied by current active research mathematicians, and the question list was assembled entirely by academics having no connection to AI companies.

\subsection{\aletheia{}'s results on \fp{}} We summarize \aletheia{}'s performance on FirstProof from \cite{feng2026aletheiatacklesfirstproofautonomously}.

We conducted two runs of \aletheia{} using two different base models: one utilized the base for model for Gemini 3 Deep Think \citep{gemini3deepthink2026} (as of February 2026), and the other with the January 2026 Gemini base model referenced in \S \ref{sec:aletheia}. For both runs, we prompted \aletheia{} with the problem statements from the \fp{} \LaTeX{} file, copy-pasted without any modification. The outputs of \aletheia{} were filtered, without any intermediate alteration, through a pre-determined verification and extraction prompt (exposed in Appendix A of \cite{feng2026aletheiatacklesfirstproofautonomously}) designed to meet the stated standards of the \fp{} authors. Moreover, the verification and extraction prompt elicited \LaTeX{} code directly as output, ensuring that manual intervention would not be required even for reformatting the response in a \LaTeX{} document. The best-of-2 evaluations for each problem, according to a majority of expert opinions, are displayed in Table \ref{table:intro-results}. 

\begin{table}[!h]
\centering
\renewcommand{\arraystretch}{1.2} 
\begin{tabular}{|l|c|c|}
\hline
    & \shortstack{\aletheia{} \\ (best of 2)}  & \shortstack{Expert Evaluation \\ (correct/total)}\\ \hline
P1  & No Output  & \\ \hline
P2  & \color{green} Correct & 4/4 \\ \hline
P3  & No Output &  \\ \hline
P4  & No Output & \\ \hline
P5  & \color{green} Correct & 4/4 \\ \hline
P6  & No Output & \\ \hline
P7  & \color{green} Correct &  3/3\\ \hline
P8  & \color{green!70} Correct? & 5/7\\ \hline
P9  & \color{green} Correct & 4/4  \\  \hline
P10 & \color{green} Correct & 2/2 \\   \hline
\end{tabular}
\caption{Summary of \aletheia{}'s performance on \fp{}. The Expert Evaluation column displays the number of experts who rated the solution as being Correct, out of the total number of experts consulted. Only the assessment on P8 was not unanimous.}
\label{table:intro-results}
\end{table}

While most of the FirstProof problems were described as Lemmas, Problem 7 was advertised as an open problem in the book \citep{Wein23}, prior to its resolution by Cappell--Weinberger--Yan (not published until the \fp{} solutions). If it had not been for the earlier solution by Cappell--Weinberger--Yan, then we would have considered this solution worthy of publication; as is, the solution is distinct enough to merit documentation in separate ArXiv paper \citepalias{FK26}.   

More details on the methodology, the results of the individual results, and the expert evaluations may be found in \cite{feng2026aletheiatacklesfirstproofautonomously}.

\subsection{Comparisons} To contextualize these results: the FirstProof authors revealed that GPT 5.2 Pro had already solved Problem 9 and Problem 10 ``out of the box'' during their internal tests. It was also observed by David Woodruff and Aryan Mokhtari that Gemini 3 Deep Think could solve Problem 10 ``out of the box''. These were the best performances by any publicly available models, and may be taken as a baseline for evaluating specialized mathematics agents. 

OpenAI claimed ``highly likely'' solutions to Problems 2, 4, 5, 6, 9, 10 using an internal model \cite{openai2026firstproof}, although their solution to Problem 2 was quickly found to be flawed; our understanding is that experts regard the other five solutions to be correct. We also note that, by self-admission, the solutions were produced with undisclosed human guidance.\footnote{See \url{https://x.com/merettm/status/2022925024798372262.}}

Additionally, Cursor researchers (Zhang and Lin) exhibited an autonomously generated solution to Problem 6; our understanding is that experts regard it to be correct. 

Aside from these instances, \emph{we are unaware of any credible claims of autonomous solutions beyond the baseline} (Problems 9 and 10),\footnote{A Zulip discussion of solutions is hosted at \url{https://icarm.zulipchat.com/\#narrow/channel/568090-first-proof/}.} despite attempts from many parties.

\subsection{Inference computation}
\begin{figure}[htbp]
    \centering
    \includegraphics[width=0.8\textwidth]{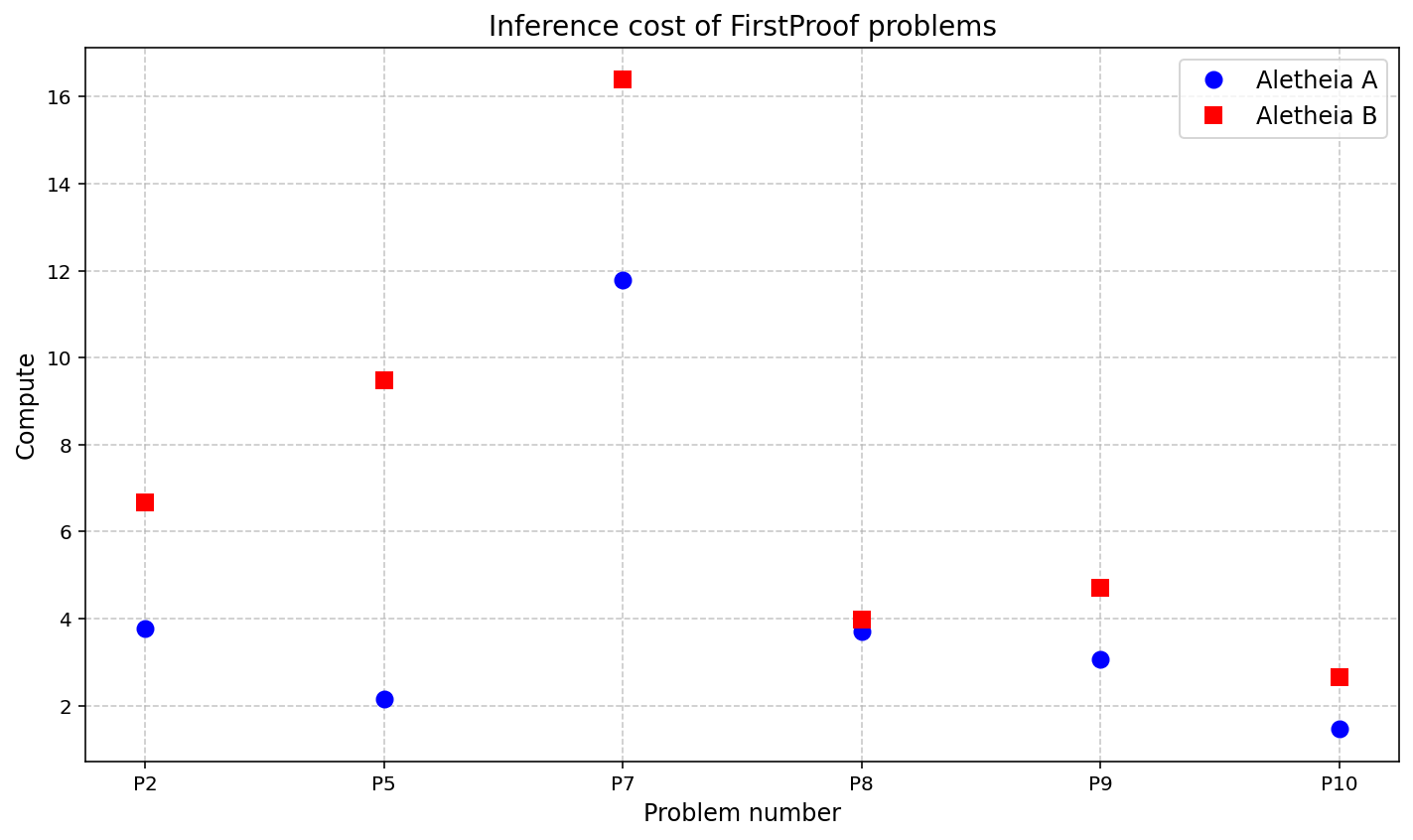}
    \caption{Plot of the inference cost per \fp{} problem, as a multiple of the inference cost of the solution to Erd\H{o}s-1051 from \citep{erdos}.}
    \label{fig:fp_compute}
\end{figure}

\aletheia{}'s inference time computation is dynamic, and can be interpreted as a rough proxy of the problem difficulty from the agent's perspective. In Figure \ref{fig:fp_compute}, we display the inference cost of each candidate solution as a multiple of the inference cost of the solution to Erd\H{o}s-1051 from \cite{erdos}. Both base models used here are different from the one used in \cite{erdos}, so the comparison is not on equal footing, but it gives some indication. For each problem, the inference cost exceeded that of Erd\H{o}s-1051. For P7 in particular, the inference cost exceeded previously observed scales by an order of magnitude. 

\section{Analysis and Discussion}

\subsection{Ablation studies}
To contextualize \aletheia{}'s performance, we conducted ablation studies using the Gemini Deep Think agent, operating at the IMO Gold scale with the identical underlying base model. We evaluated Deep Think using the same prompts that generated the research results detailed in \S \ref{sec:math-results}.

 We first evaluate the 13 Erd\H{o}s problems that \aletheia{} successfully solved, as listed in \S \ref{ssec:erdos}. According to human expert grading\footnote{Natural language verification by human experts inherently involves some subjectivity; see Footnote 3 of \cite{erdos}.}, Gemini Deep Think correctly solved 8 of these 13 problems, while operating at almost exactly twice the average compute per problem as \aletheia{}. These results are summarized in Table \ref{table:ablation-erdos}. 

\begin{table}[h]
\centering
\setlength{\tabcolsep}{4pt} % Adjusts spacing to fit all 13 columns nicely
\begin{tabular}{|*{13}{c|}} 
\hline 
 333 & 397 & 591 & 652 & 654 & 659 & 705 & 935 & 992 & 1040 & 1051 & 1089 & 1105 \\ \hline
 {\color{green}\ding{51}} & {\color{green}\ding{51}} & {\color{red}\ding{55}} & {\color{green}\ding{51}} & {\color{green}\ding{51}} & {\color{green}\ding{51}} & {\color{red}\ding{55}} & {\color{green}\ding{51}} & {\color{red}\ding{55}} & {\color{red}\ding{55}} & {\color{green}\ding{51}} & {\color{green}\ding{51}} & {\color{red}\ding{55}} \\ \hline
\end{tabular}
\caption{Performance of Gemini Deep Think (IMO scale) on the 13 Erd\H{o}s problems successfully solved by \aletheia{} (\S \ref{ssec:erdos}). Both agents utilized the same base model, with Deep Think consuming roughly 2$\times$ the average compute per problem.}\label{table:ablation-erdos}
\end{table}

Next we compare performance on the specific prompts underlying the research papers documented in \S \ref{sec:math-results}. As summarized in Table \ref{table:ablation-papers}, Deep Think successfully reproduced the correct answer for \citepalias{FYZ4} but failed across all three prompts for \citepalias{Feng2026}. For the prompts in \citepalias{LeeSeo2026}, it resolved the first prompt but failed on the second and far more crucial prompt. For \citepalias{BKKKZ}, the interaction was too messy to replicate exactly, but Deep Think essentially succeeded. Lastly, for \citepalias{ACGKMP}, Deep Think derived a satisfactory upper bound, although it was not as sharp as \aletheia{}'s. On this set of prompts, the total compute used by the two agents turned out to be comparable. The results are summarized in Table \ref{table:ablation-papers}.

\begin{table}[h]
\centering
\setlength{\tabcolsep}{4pt} % Adjusts spacing to fit all 13 columns nicely
\begin{tabular}{|*{5}{c|}} 
\hline 
 \citepalias{FYZ4} & \citepalias{Feng2026} & \citepalias{LeeSeo2026} & \citepalias{BKKKZ} & \citepalias{ACGKMP}  \\ \hline
 {\color{green}\ding{51}} &  {\color{red}\ding{55}} & {\color{red}\ding{55}} & {\color{green}\ding{51}}  & {\color{green}\ding{51}}  \\ \hline
\end{tabular}
\caption{Success of Gemini Deep Think (IMO scale) at reproducing the autonomous assistance utilized for our research papers. Both agents utilized the same base model and consumed similar total compute.}\label{table:ablation-papers}
\end{table}

\subsection{Accuracy} The results of this paper should \textbf{not} be interpreted as suggesting that AI can consistently solve research-level mathematics questions. In fact, our anecdotal experience is the opposite: success cases are rare, and an apt intuition for autonomous capabilities (and limitations) may currently be important for finding such cases. The papers \citepalias{Feng2026, LeeSeo2026, ACGKMP} grew out of spontaneous positive outcomes in a wider benchmarking effort on research-level problems; for most of these problems, no autonomous progress was made.  

\textbf{Erd\H{o}s problems.} The case study \cite{erdos} on the Erd\H{o}s problems offers a more quantitative picture. Table \ref{tab:solution-candidates} illustrates the success rate on the 200 solution candidates that we were able to mark Correct or Incorrect (filtered from 700 open problems). In total, 31.5\% of the solutions were technically correct under some interpretation of the question, but only 6.5\% were meaningfully correct in addressing what we deemed to be the \emph{intended} interpretation. 

\begin{table}[h]
\centering
\begin{tabular}{lrr}
\toprule
\textbf{Category} & \textbf{Count} & \textbf{Percentage} \\
\midrule
Fundamentally Flawed & 137 & 68.5\% \\
Technically Correct  & 63  & 31.5\% \\
\hspace{1em} \textit{Meaningfully Correct (subset)} & \textit{13} & \textit{6.5\%} \\
\midrule
\textbf{Total Candidates} & \textbf{200} & \textbf{100.0\%} \\
\bottomrule
\end{tabular}
\caption{Solution accuracy on 200 AI-generated responses, as graded by humans.}
\label{tab:solution-candidates}
\end{table}
We note that this study was doing with the November 2025 base model of Gemini 3, which turned out to be substantially weaker than the January 2026 base model used in Section \ref{sec:aletheia} and for \fp{}.

\textbf{FirstProof.} On the 10 \fp{} problems, both of our runs on \aletheia{} produced solution candidates to exactly 6 problems (P2, P5, P7, P8, P9, P10). From a best-of-2 evaluation, the majority opinion of expert evaluations indicated that all 6 problems were solved correctly (under the interpretation of being publishable after minor revisions), although the assessments on P8 were not unanimous: only 5 out of 7 experts rated it Correct. The assessment of individual solutions is displayed in Table \ref{table:resultsafter}.

\begin{table}[h]
\centering
\renewcommand{\arraystretch}{1.5} % Adjusts row height for readability
\begin{tabular}{|l|l|l|l|}
\hline
   & \textbf{Run A} &  \textbf{Run B} & \textbf{Zulip}\\ \hline
P1 & No output &  No output & \\ \hline
P2 & \color{green} Correct &  \color{green} Correct  &  \href{https://icarm.zulipchat.com/#narrow/channel/568090-first-proof/topic/Problem.202.20--.20Aletheia/with/574567015}{Link}\\ \hline
P3 & No output & No output &  \\ \hline
P4 & No output &  No output & \\ \hline
P5 & \color{green} Correct & {\color{red}Misinterpreted} & \href{https://icarm.zulipchat.com/#narrow/channel/568090-first-proof/topic/Problem.205.20--.20Aletheia/with/575042104}{Link} \\ \hline
P6 & No output &  No output & \\ \hline
P7 & \color{red}Critically Flawed &  \color{green} Correct & \href{https://icarm.zulipchat.com/#narrow/channel/568090-first-proof/topic/Problem.207.20--.20Aletheia/with/574990987}{Link} \\ \hline
P8 & \color{red} Inadequate   &  \color{green!70}Correct?  & \href{https://icarm.zulipchat.com/#narrow/channel/568090-first-proof/topic/Problem.208.20--.20Aletheia/with/574569368}{Link} \\ \hline
P9 & \color{green} Correct &  \color{green} Correct & \href{https://icarm.zulipchat.com/#narrow/channel/568090-first-proof/topic/Problem.209.20--.20Aletheia/with/574726804}{Link} \\ \hline
P10 & \color{green} Correct & \color{green} Correct & \href{https://icarm.zulipchat.com/#narrow/channel/568090-first-proof/topic/Problem.2010.20--.20Aletheia/with/574570445}{Link} \\ \hline
\end{tabular}
\caption{Evaluation of the results from the two runs of \aletheia{}, according to a majority of expert assessments. On Problem 8, the expert assessment was not unanimous, with 5/7 rating the solution as Correct. We include links with public comments (on Zulip) for individual problems. 
}
\label{table:resultsafter}
\end{table}

For the other 4 problems (P1, P3, P4, P6) both of our agents returned no solution: either by explicitly outputting ``No solution found'', or by not returning any output within the time limit. The overall accuracy, as well as conditional accuracy on answered problems, was consistent with our observations on \emph{FutureMath} Basic, the internal benchmark discussed in \S \ref{sec:aletheia} and aimed at a comparable level of material. 

We consider one of \aletheia{}'s correct \fp{} solutions to be publication-grade (Problem 7). This success rate (1/10) for publication-grade results is markedly lower than the success rate for producing the results documented in \S \ref{sec:math-results}. We note that most of the \fp{} problems are intended as technical Lemmas, not publication-worthy as standalone results.

\subsection{Weaknesses of AI}
We identify several qualitative gaps between current autonomous mathematics research and human-generated research. To date, autonomous results have been relatively brief and elementary in comparison to typical human papers. %This is likely because the potential complexity of responses from natural language models is limited by context constraints. 
Furthermore, success cases seem to arise from clever technical manipulations or vast knowledge retrieval, rather than what mathematicians would consider to be genuine creativity, although the latter concept is admittedly subjective. %This may stem from a fundamental tension: generative models are designed to replicate an existing distribution, whereas the essence of research is the discovery of out-of-distribution ideas.

Even with its verifier mechanism, \aletheia{} is still more prone to errors than human experts. Furthermore, whenever there is room for ambiguity, the model exhibits a tendency to misinterpret the question in a way that is easiest to answer, even when such an interpretation would be obviously unintended to a human expert. This aligns with the well-known tendencies for ``specification gaming'' and ``reward hacking'' in machine learning. Finally, we reiterate that hallucination is still a common failure mode. Even with internet search capability to check references, the model tends to fabricate or misrepresent results from legitimate references in order to assert a solution (see Figure \ref{fig:hallucinated-theorem} for an example).

\section{Representing AI contributions to mathematics} 
Mathematics provides a domain that is---at least superficially---analogous to game-playing, making it a highly attractive target for AI researchers. In mathematics there is a near-universal consensus on starting points (mathematical axioms) and atomic moves (valid mathematical deductions), and a rich landscape of objectives (precise theorems and conjectures). 
However, the extreme difficulty and specialization of frontier mathematics research makes performance evaluation exceptionally challenging. For a given research problem, there may be only a handful of mathematicians in the world equipped to evaluate a potential solution, and such evaluations can take tremendous effort. (For example, in mathematics the peer review process typically exceeds a year, and it is common for research papers to surpass 100 pages in length.) Even given a solution that is known to be correct, perhaps through formal verification, it could be difficult for an outside observer to distinguish a major advance from a triviality. 

This evaluation gap has enabled a surge in hyperbolic or misleading messaging regarding AI-generated mathematical content. Instead of using academic publishing venues with vetting and review by professional mathematicians, such claims are often propagated through social media channels. While AI proponents have obvious incentives to exaggerate the mathematical capabilities of AI, a somewhat perverse incentive has also arisen for mathematicians to overstate the contribution of AI to their own work, because doing so leads to a significant increase in publicity and attention. Moreover, misinformation propagates easily in this domain because the audience is ill-equipped to evaluate the claims.

Since we are developing tools for AI-assisted mathematics research, we also felt a responsibility to contribute to the discussion on responsible documentation, evaluation and communication of AI-generated results. In order to inform our thinking, we held a discussion with several mathematicians to understand the concerns of the mathematical community regarding AI.\footnote{Mathematicians consulted (outside those among the authors of this paper) included Jarod Alper$^\star$, Kevin Barreto, Sourav Chatterjee, Otis Chodosh, Michael Hutchings, Daniel Litt, Ciprian Manolescu, Mona Merling, Johannes Schmitt, Ravi Vakil, Shengtong Zhang. $^\star$Affiliated with Google DeepMind.} One of the dominant themes of the discussion was the desire for \emph{transparency}, both about the role of the AI and the technical importance of AI-assisted results. Motivated by this, we will define \emph{one} possible framework for describing autonomous mathematics, and contextualize our own results within it for illustration. This should be considered as only a starting point in a broader discussion of responsible scientific communication.  If any such standards are to be universally adopted, they must arise organically from the mathematical community. 

\subsection{Autonomous Mathematics Levels}\label{ssec:AMLevels}

Autonomous driving is classified according to standard ``Levels'', measuring the degree of autonomy, defined by the Society of Automotive Engineers (SAE). A meaningful taxonomy of autonomous mathematics should have at least \emph{two dimensions}, with one axis for the \textbf{level of autonomy} and another for the \textbf{level of mathematical significance}\footnote{A similar example of such a division is in Epoch AI's FrontierMath open problems, which are rated as ``moderately interesting,'' ``solid result,'' ``major advance,'' or ``breakthrough.'' \cite{epochai:2026:frontiermathopen}}. Maintaining a coarse division while preserving essential distinctions would minimize subjective disputes. We divide our own results into three broad levels of autonomy, described in Table \ref{tab:axis1}. 

\begin{table}[h]
\centering
\small
\renewcommand{\arraystretch}{1.6} 
\begin{tabular}{|>{\centering\arraybackslash\bfseries}m{1cm}|>{\RaggedRight\arraybackslash\bfseries}m{3cm}|>{\RaggedRight\arraybackslash}m{10cm}|}
\hline
\rowcolor{tablegray} \textbf{Level} & \textbf{Level} & \textbf{Description} \\ \hline
H & Primarily Human \newline (secondary AI input) & Core mathematical content is human-generated; AI contributions are minor or auxiliary (e.g., literature search, basic calculations, writing routine proofs which the human author could have supplied) if any. \\ \hline
C & Human-AI\newline Collaboration & A substantive human-AI collaboration where both parties contribute in an essential way. \\ \hline
A & Essentially\newline Autonomous & The core mathematical content is fully AI-generated without any essential human intervention. (Human authors are still ultimately responsible for the final paper content.) \\ \hline
\end{tabular}
\caption{Axis 1: Level of Autonomy. This axis quantifies the relative contribution between the human and AI agents.}
\label{tab:axis1}
\end{table}

\subsubsection{Discussion and Examples for Level of Autonomy.} In these terms, we consider the paper \citepalias{Feng2026}  to be of Level A, as the human contribution is only in expository aspects, aside from posing the original questions. The results in \citep{erdos} are also of Level A insofar as the role of humans was only to correct minor inaccuracies. We consider the papers \citepalias{LeeSeo2026} and \citepalias{BKKKZ} to be of Level C: both humans and AI contributed essential aspects of the papers. We consider the papers \citepalias{FYZ4} and \citepalias{ACGKMP} to be of Level H: while the written papers have non-trivial contributions from AI, these contributions form a minor portion of the paper and the human authors could have supplied those parts themselves.

Even with these broad subdivisions, there can be some ambiguity. A possible situation is that interacting with AI leads human mathematicians to a key idea or reference, but the AI itself does not truly make its own mathematical contribution. In other words, the AI contributes only via ``inspiration''. Although the research may not have happened without AI assistance, we would consider such instances as Level H.

\subsubsection{Discussion and Examples for Level of Significance.} We emphasize at the outset that these Levels fail to capture the depth and diversity of human mathematics research, and are only meant to contextualize autonomous contributions. Given the inherent subjectivity of mathematical significance, and the proven difficulty of comparing the work of (human) mathematicians, we felt that the Levels of Mathematical Significance should be coarse enough that they do not differentiate human-generated research. A possible exception is that there is near-universal consensus on the top 5 journals in mathematics\footnote{In alphabetical order, these are: \textit{Acta Math., Annals of Math., Inventiones Math., Journal of the AMS, Publ. Math. IHES}. We expect that any proposed list of ``top $N$ journals'' for a different value of $N$ would be met with significant contention, as would any attempted internal ranking of these five.}, so it could make sense to anchor one level to them. One would expect to find major advances published in these journals; as far as we are aware, no autonomous result has yet come close to that standard. A final category could be reserved for truly landmark breakthroughs of once-in-a-generation stature.\footnote{Examples might include the proof of \emph{Fermat's Last Theorem} \cite{wiles1995modular, taylor1995ring} or Perelman's proof of the \emph{Poincar\'e Conjecture} \cite{perelman2002entropy, perelman2003ricci, perelman2003finite}.} Although evident, it seems worth emphasizing that \emph{the Level of Mathematical Significance can only be evaluated by mathematicians, who are moreover experts on the particular domain of the problem}. 

\begin{table}[h]
\centering
\small
\renewcommand{\arraystretch}{1.6} 
% Total width is roughly 14.5cm to fit standard A4/Letter margins
% m{} ensures vertical centering
\begin{tabular}{|>{\centering\arraybackslash}m{1cm}|>{\RaggedRight\arraybackslash\bfseries}m{3.0cm}|>{\RaggedRight\arraybackslash}m{10cm}|}
\hline
\rowcolor{tablegray} \textbf{Level} & \textbf{Significance} & \textbf{Examples / Criteria} \\ \hline
0 & Negligible novelty & Suitable for Olympiad contests or Ph.D. student exercises. \\ \hline
1 & Minor novelty & A new result, but the techniques and novelty do not meet the standard for a professional research paper. \\ \hline
2 & Publication Grade & Publishable in a research journal in good standing (that is not specifically intended for students or expository articles), following established peer-review practices. \\ \hline
3 & Major\newline Advance & At the highest level of typical human results; published in the top 5 general-interest journals in mathematics. \\ \hline
4 & Landmark\newline Breakthrough & A truly landmark breakthrough result of generational importance and influence.  \\ \hline
\end{tabular}
\caption{Axis 2: Mathematical Significance of Result. This axis measures how significant the mathematical result is, regardless of how it was produced.}
\label{tab:axis2}
\end{table}

 In these terms, we view our solutions to Erd\H{o}s-652, 654, 935, 1040 as Level A0, and our solution to Erd\H{o}s-1051 as Level A1. We view \citepalias{Feng2026} as Level A2, \citepalias{BKKKZ} and \citepalias{LeeSeo2026} as Level C2, and \citepalias{ACGKMP} and \citepalias{FYZ4} as Level H2. We reiterate that the variation within Level 2 is deliberately wide to encompass the vast majority of papers written by professional mathematicians, ranging from very minor to extremely influential works. This appears to be the only way to avoid subjective disputes about human research. Because of the exceptional width of this category, \textbf{the public should not understand a paper of level C2 or A2 as indicating that AI tools have reached the level of human mathematicians.} For example, we view our own results within Level 2 \citepalias{ACGKMP, BKKKZ, Feng2026, FYZ4, LeeSeo2026} as being very spread out in significance and difficulty.

\subsection{Documenting Human-AI Collaboration}\label{ssec:human-AI} Another theme of concern in the discussion with mathematicians was how best to document AI-assisted mathematics, both in the interest of helping the community achieve effective usage of AI tools, and also in regulating the promotion of AI involvement as an attention-getting mechanism. Several examples of AI-assisted mathematics papers have recently appeared in the literature \citepalias{bryan2026motivic, jang2025point, schmitt2025extremal}, with each featuring its own method of documenting the AI contributions. In our work, the interactions were clean enough to be neatly captured by a ``Human-AI Interaction (HAI) Card'', which we display below for the papers \citepalias{Feng2026, LeeSeo2026, BKKKZ}. This terminology is inspired by the ``model card'' concept in describing AI models.

\begin{interactionlog}[\citepalias{Feng2026}]{https://github.com/google-deepmind/superhuman/tree/main/aletheia}

    \human{Query the computation of eigenweights (certain structure constants arising in arithmetic geometry) for groups of Type A.}

    \ai{\aletheia}{Fully correct solution using Atiyah--Bott localization, Schur polynomial manipulation, Frobenius character identities, and the Murnaghan--Nakayama rule.}

    \human{Proceed to query eigenweights for groups of Type C.}

    \ai{\aletheia}{Fully correct solution, applying the same tools in a variant manner.}

    \human{Proceed to query eigenweights for groups of Type D.}

    \ai{\aletheia}{Fully correct solution, applying the same tools in a variant manner.}

\end{interactionlog}

\begin{interactionlog}[\citepalias{LeeSeo2026}]{https://github.com/google-deepmind/superhuman/tree/main/aletheia}
    
    \human{Query a multivariate inequality on independent sets of graphs (generalization of \cite{sah2019number}).}
    
    \ai{Gemini\,3 Deep Think}{Fully correct solution.}

    \human{Query challenging extension to semiproper colourings.}
    
    \ai{\aletheia}{Proof outline with crucial ideas, including dual sets and their log convexity, reduction techniques, and several key Lemmas. }

    \human{Work out details of outline, devising new (simpler) proofs.}

\end{interactionlog}

\begin{interactionlog}[\citepalias{BKKKZ}]{https://github.com/google-deepmind/superhuman/tree/main/aletheia}
    
    \human{Query Erd\H{o}s-1051 as stated in~\citep{ErdosGraham1980}, involving summation of $a_n a_{n+1}$ case.}
    
    \ai{\aletheia}{Correct solution.}

    \human{State and prove a generalization to the $a_n\cdots a_{n+d-1}$ case, and requests a ``meaningful, interesting generalization''.}
    
    \ai{Gemini\,3 Deep Think}{Propose a generalization to the $a_n^{w_0}\cdots a_{n+d-1}^{w_{d-1}}$ case; provides non-rigorous heuristics.}

    \human{Further generalize the theorem by changing the hypothesis $\lim=\infty$ to a weaker condition $\limsup=\infty$; complete a rigorous proof; identify examples demonstrating that the results are optimal.}

\end{interactionlog}

In other instances, it may not be realistic to capture the interactions with the same template.  While the specific presentation may be left to an author's discretion, the field would benefit from a common baseline for transparency. For Level C or Level A results, where AI input is deemed essential, a possible baseline would be to expose at least the most important raw prompts and outputs that contain the essential new insights generated by AI. Specific standards are left for the community of mathematicians to decide. We note that, in a related direction, Tao has initiated an online discussion of ``best practices for incorporating AI usage'' \cite{tao2024bestpractices} for mathematicians.

\section{Reflections on the Impact of AI in Mathematics}

To date, hype notwithstanding, the impact of artificial intelligence on pure mathematics research has been limited. While our results do solve some problems that seem to have eluded experts, they do not indicate that artificial intelligence has matched, or will match, the capabilities of human mathematicians. Rather, they illustrate how certain comparative advantages of AI models over humans can be useful for certain kinds of problems. This perhaps clarifies the directions where human researchers can expect the most impact from AI in the near future.

A first observation is that AI models exhibit a form of intelligence that diverges significantly from that of human scientists. In any specific subject, frontier models have much shallower knowledge than a domain expert, but they also possess superhuman \emph{breadth} of knowledge, which could be the key to unlocking certain problems. The simple fact that artificial intelligence differs from human intelligence presents the possibility that it is better suited for solving some types of problems, for example those requiring vast memory, computation, or breadth of knowledge.

Another comparative strength of AI is that it is not constrained by human physical limitations. It is likely that many open questions lie within the reach of existing techniques, but are not resolved because of limited time and attention from the right experts, as demonstrated by our results on the Erd\H{o}s problems \cite{erdos}. This reinforces the point that AI is bottlenecked by very different factors compared to humans, which can be an advantage in the right context.

\section{Related Work}

\paragraph{Advances in Mathematical Reasoning}
Recent advances in large language models have yielded systems capable of sophisticated mathematical reasoning \cite{deepmind:gemini2p5deepthink, openai:2025:gpt5_2}, moving beyond standard benchmarks to competition-level problems. Notably, these models have achieved gold-medal standards at the International Mathematics Olympiad (IMO) \cite{imo-gold} and demonstrated proficiency on difficult benchmarks like \textit{IMO-ProofBench} \cite{luong2025robustmathematicalreasoning}. New, more challenging benchmarks continue to emerge as the field shifts to focus on research-level math \cite{glazer2024frontiermath,nie2025uqassessinglanguagemodels, tao2026erdosai}.

\paragraph{AI-Assisted Research Results}
Already, AI has been successfully applied to assist mathematical research. Examples include research on extremal descendant integrals on moduli spaces \citepalias{schmitt2025extremal}, motivic classes of genus zero maps \citepalias{bryan2026motivic}, and convergence rates for Nesterov's method \citepalias{jang2025point}. Beyond these specific cases, AI has been used in other works across discrete analysis, convex analysis, and combinatorics \citepalias{ivanisvili2025counterexamplemajorityoptimalitynicd, salim2025acceleratingmathematicalresearchlanguage}. Other studies show AI is useful for solving accessible conjectures or assisting with problems that require mathematical exploration \citepalias{feldman2025godeltestlargelanguage, georgiev2025mathematicalexplorationdiscoveryscale}. Even more broadly, \cite{woodruff2026acceleratingscientificresearchgemini} demonstrates a multitude of techniques and case studies for AI-assisted research across several mathematics-adjacent domains, including theoretical computer science, economics, and physics.

\paragraph{Frameworks for Human-AI Collaboration}
A variety of human-AI collaborative implementations have emerged, ranging from fully agentic architectures with self-verification and refinement \cite{huang2025winninggoldimo2025, bryan2026motivic}, to exploration methods such as AlphaEvolve \cite{novikov2025alphaevolvecodingagentscientific, wang2025thetaevolvetesttimelearningopen, yuksekgonul2026learningdiscovertesttime}, to more collaboration-focused methods such as interactive protocols suggesting how to integrate AI as a research assistant 
\cite{li2025advancingmathematicalresearchhumanai,henkel2025mathematiciansassistantintegratingai}. Despite the surge in attention \cite{bubeck2025earlyscienceaccelerationexperiments,zheng2025automationautonomysurveylarge}, the community is still in the early stages of adapting AI for mathematical research, making it essential to establish transparent standards for how AI-generated mathematics is documented, evaluated, and communicated.

\section{Conclusion}

Ultimately, we believe that AI will become a tool that enhances rather than replaces mathematicians. Currently, natural language models struggle to reason reliably without human intervention to correct mistakes and hallucinations, while formal verification systems are not yet capable of even formulating the questions of interest on most research frontiers. For this reason, we have introduced specialized math reasoning agents, incorporating informal natural language verification, to help mathematicians harness the benefits of AI.

\paragraph{Acknowledgments.} Special thanks to the following experts for mathematical discussions on the work: Jarod Alper, Kevin Barreto, Thomas Bloom, Sourav Chatterjee, Otis Chodosh, Michael Hutchings, Seongbin Jeon, Youngbeom Jin, Aiden Yuchan Jung, Jiwon Kang, Jimin Kim, Vjekoslav Kovač, Daniel Litt, Ciprian Manolescu, Mona Merling, Agustin Moreno, Carl Schildkraut, Johannes Schmitt, Insuk Seo, Jaehyeon Seo, Cheng-Chiang Tsai, Ravi Vakil, Zhiwei Yun, Shengtong Zhang, Wei Zhang, Yufei Zhao.
We also would like to thank Adam Brown, Alex Davies, Thomas Hubert, Pushmeet Kohli, Vinay Ramasesh, Eugénie Rives, Benoit Schillings, Divy Thakkar for feedback and support of the work.

\bibliography{arxiv_latest}

\appendix

\newpage

\onecolumn
\section{Case Study: FutureMath Basic}\label{FMbasic}

As mentioned in the main text, we curated an internal benchmark of of advanced (Ph.D. level) exercises for evaluation of \aletheia{}. We give one example of a FutureMath Basic problem solved by \aletheia{}. Gemini 3 Deep Think made negligible progress on this problem. 

\noindent\textbf{FM-Grad-011:}
\begin{quote}
    Let a positive integer $q \ge 3$ be given, and define $S = \{1,2,\dots,q\}$.
For a positive integer $N \ge 2$ and $x = (x_1,\dots,x_N) \in S^N$, define
\[
U_N(x) = \frac{1}{2N} \sum_{1 \le i,j \le N} \mathbf{1}(x_i = x_j).
\]

For a real number $\beta > 0$, define
\[
Z_{N,\beta} = \sum_{x=(x_1,\dots,x_N) \in S^N} e^{\beta U_N(x)}.
\]

Find the maximum value of $\beta$ for which the following equality holds:
\[
\lim_{N \to \infty} \frac{1}{N} \log Z_{N,\beta}
= \frac{\beta}{2q} + \log q.
\]
\end{quote}

\noindent\textbf{Short answer:} $\frac{2(q-1)}{q-2} \log (q-1)$

\noindent\textbf{Aletheia solution:}
\begin{quote}
    
\textbf{Large Deviation Analysis of the Phase Transition in the Mean-Field Potts Model}

\textit{
We determine the maximum inverse temperature $\beta$ for which the free energy density of the $q$-state mean-field Potts model is determined by the uniform empirical distribution. Using large deviation theory, we rigorously reduce the optimization of the free energy functional to a one-dimensional problem and solve the resulting system of algebraic equations to identify the exact critical temperature of the first-order phase transition.
}

\textbf{Introduction and Formulation via Empirical Measures}

Let $S = \{1, 2, \dots, q\}$ for an integer $q \ge 3$. We consider the partition function
\[
Z_{N,\beta} = \sum_{x \in S^N} e^{\beta U_N(x)},
\]
where the energy function is given by $U_N(x) = \frac{1}{2N} \sum_{i,j=1}^N \mathbf{1}(x_i = x_j)$. 

For a configuration $x = (x_1, \dots, x_N) \in S^N$, let $N_k(x) = \sum_{i=1}^N \mathbf{1}(x_i = k)$ denote the occupation number of state $k$. The empirical probability measure of the configuration $x$ is the vector $p(x) = (p_1(x), \dots, p_q(x)) \in \Delta_q$, where $p_k(x) = N_k(x)/N$ and $\Delta_q = \{ p \in \mathbb{R}^q : p_k \ge 0, \sum_k p_k = 1 \}$ is the $(q-1)$-dimensional probability simplex. The energy can be expressed purely in terms of the empirical measure:
\[
U_N(x) = \frac{N}{2} \sum_{k=1}^q p_k(x)^2.
\]

By Sanov's Theorem (see Ellis, 1985), the sequence of empirical measures satisfies a large deviation principle with rate function given by the negative of the Shannon entropy, $H(p) = -\sum_{k=1}^q p_k \log p_k$. Applying Varadhan's Lemma, the asymptotic free energy density is the supremum over the simplex of the difference between the energy and the entropy:
\begin{equation} \label{eq:varadhan}
\lim_{N \to \infty} \frac{1}{N} \log Z_{N,\beta} = \sup_{p \in \Delta_q} F_\beta(p),
\end{equation}
where the free energy functional is
\[
F_\beta(p) = \frac{\beta}{2} \sum_{k=1}^q p_k^2 - \sum_{k=1}^q p_k \log p_k.
\]

\textbf{The Uniform Distribution and Symmetry Breaking}

At the uniform distribution $p^* = (1/q, 1/q, \dots, 1/q)$, the value of the functional is
\[
F_\beta(p^*) = \frac{\beta}{2} \sum_{k=1}^q \frac{1}{q^2} - \sum_{k=1}^q \frac{1}{q} \log \frac{1}{q} = \frac{\beta}{2q} + \log q.
\]
The equality stated in the problem holds if and only if $p^*$ is the global maximizer of $F_\beta(p)$ on $\Delta_q$.

To identify all possible local maximizers of $F_\beta$, we optimize $F_\beta(p)$ subject to $\sum_{k=1}^q p_k = 1$. Using a Lagrange multiplier $\lambda$, the critical points in the interior of $\Delta_q$ satisfy the stationarity condition
\[
\beta p_k - \log p_k - 1 = \lambda \quad \text{for all } k = 1, \dots, q.
\]
Define the function $f(z) = \beta z - \log z$. Since $f''(z) = z^{-2} > 0$, $f(z)$ is strictly convex on $(0, 1)$. Thus, any horizontal line $f(z) = \lambda + 1$ can intersect the graph of $f(z)$ at most twice. This implies that the components of any critical point $p$ can take at most two distinct values.

Consequently, any critical point must be of the form $p = (x, y, \dots, y)$ up to a permutation of the coordinates, where $x$ appears $m$ times and $y$ appears $q-m$ times. A standard second-order stability analysis of the Hessian of $F_\beta$ (Costeniuc, Ellis, and Touchette, 2005) reveals that local maxima can only occur when $m=1$ or $m=q$. Therefore, apart from the uniform distribution $p^*$, the only candidate for a global maximum is a distribution with one large component and $q-1$ small components: $p = (x, y, y, \dots, y)$ with $x + (q-1)y = 1$.

Eliminating $y = \frac{1-x}{q-1}$, the objective function restricted to this symmetric subspace becomes a function of $x \in (0, 1)$:
\[
h(x) = \frac{\beta}{2} \left( x^2 + \frac{(1-x)^2}{q-1} \right) - x \log x - (1-x) \log \left( \frac{1-x}{q-1} \right).
\]

\textbf{Determination of the Critical Inverse Temperature}

For $q \ge 3$, the model undergoes a first-order phase transition (Wu, 1982). As $\beta$ increases, a secondary local maximum appears at some $x > 1/q$ and eventually overtakes the uniform state. The critical inverse temperature $\beta_{\max}$ is defined as the point at which this secondary peak matches the height of the uniform state. This requires finding $x_c \in (1/q, 1)$ and $\beta_{\max}$ such that:
\begin{enumerate}
    \item \textbf{Stationarity:} $h'(x_c) = 0$
    \item \textbf{Equal heights:} $h(x_c) = h(1/q)$
\end{enumerate}

The derivative of $h(x)$ is
\[
h'(x) = \beta \left( \frac{qx-1}{q-1} \right) - \log \left( \frac{x(q-1)}{1-x} \right).
\]
Setting $h'(x_c) = 0$ yields
\begin{equation} \label{eq:beta}
\beta = \frac{q-1}{qx_c-1} \log \frac{x_c(q-1)}{1-x_c}.
\end{equation}

To determine $x_c$, we substitute this expression for $\beta$ into the equal-height condition $h(x_c) - h(1/q) = 0$. Using the algebraic identity $\frac{1}{2}\left(x_c^2 + \frac{(1-x_c)^2}{q-1}\right) - \frac{1}{2q} = \frac{(qx_c-1)^2}{2q(q-1)}$, the condition becomes
\[
\frac{1}{2} \left( \frac{qx_c-1}{q} \right) \log \frac{x_c(q-1)}{1-x_c} - x_c \log x_c - (1-x_c) \log \left(\frac{1-x_c}{q-1}\right) - \log q = 0.
\]
This system of equations can be solved exactly. One can verify algebraically that $x_c = \frac{q-1}{q}$ is the unique non-trivial solution in the interval $(1/q, 1]$. With $x_c = \frac{q-1}{q}$, we have $1-x_c = \frac{1}{q}$, and the argument of the logarithm in \eqref{eq:beta} becomes $(q-1)^2$. Substituting $x_c$ into \eqref{eq:beta} gives:
\[
\beta_{\max} = \frac{q-1}{q\left(\frac{q-1}{q}\right)-1} \log \left( (q-1)^2 \right) = \frac{2(q-1)}{q-2} \log(q-1).
\]

To verify the equal-height condition at this critical point, we compute $h(x_c) - h(1/q)$ with $\beta_{\max}$ and $x_c = \frac{q-1}{q}$:
\[
h(x_c) - h(1/q) = \frac{\beta_{\max}}{2} \left(\frac{q^2-3q+3}{q(q-1)} - \frac{1}{q}\right) - \frac{q-2}{q} \log(q-1).
\]
Using $\frac{q^2-3q+3}{q(q-1)} - \frac{1}{q} = \frac{(q-2)^2}{q(q-1)}$, this becomes
\[
\frac{1}{2}\left( \frac{2(q-1)}{q-2} \log(q-1) \right) \frac{(q-2)^2}{q(q-1)} - \frac{q-2}{q} \log(q-1) = 0,
\]
which holds identically.

\textbf{Conclusion}

For $\beta \le \beta_{\max}$, the uniform distribution $p^* = (1/q, \dots, 1/q)$ is the global maximizer of the free energy functional $F_\beta(p)$, and the limit of the free energy density is $\frac{\beta}{2q} + \log q$. For $\beta > \beta_{\max}$, the system undergoes a first-order phase transition; a non-uniform configuration yields a strictly larger free energy density, violating the equality. 

Thus, the maximum value is
\[
\beta_{\max} = \frac{2(q-1)}{q-2} \log(q-1).
\]

\textbf{References}
\begin{enumerate}
    \item Costeniuc, M., Ellis, R. S., and Touchette, H. (2005). Complete analysis of phase transitions and ensemble equivalence for the Curie–Weiss–Potts model. \textit{Journal of Mathematical Physics}, 46(6), 063301.
    \item Ellis, R. S. (1985). \textit{Entropy, Large Deviations, and Statistical Mechanics}. Springer-Verlag. (Theorem II.7.1, Application to the Curie-Weiss model).
    \item Wu, F. Y. (1982). The Potts model. \textit{Reviews of Modern Physics}, 54(1), 235--268.
\end{enumerate}

\end{quote}

\onecolumn
\section{Case Study: FutureMath Basic}\label{FMbasic}

This is an example of FutureMath Basic problems that Aletheia solved using the search tool. Gemini Deep Think did not make much progress on this problem.

\noindent\textbf{FM-Grad-011:}
\begin{quote}
    Let a positive integer $q \ge 3$ be given, and define $S = \{1,2,\dots,q\}$.
For a positive integer $N \ge 2$ and $x = (x_1,\dots,x_N) \in S^N$, define
\[
U_N(x) = \frac{1}{2N} \sum_{1 \le i,j \le N} \mathbf{1}(x_i = x_j).
\]

For a real number $\beta > 0$, define
\[
Z_{N,\beta} = \sum_{x=(x_1,\dots,x_N) \in S^N} e^{\beta U_N(x)}.
\]

Find the maximum value of $\beta$ for which the following equality holds:
\[
\lim_{N \to \infty} \frac{1}{N} \log Z_{N,\beta}
= \frac{\beta}{2q} + \log q.
\]
\end{quote}

\noindent\textbf{Short answer:} $\frac{2(q-1)}{q-2} \log (q-1)$

\noindent\textbf{Aletheia solution:}
\begin{quote}
    
\textbf{Large Deviation Analysis of the Phase Transition in the Mean-Field Potts Model}

\textit{
We determine the maximum inverse temperature $\beta$ for which the free energy density of the $q$-state mean-field Potts model is determined by the uniform empirical distribution. Using large deviation theory, we rigorously reduce the optimization of the free energy functional to a one-dimensional problem and solve the resulting system of algebraic equations to identify the exact critical temperature of the first-order phase transition.
}

\textbf{Introduction and Formulation via Empirical Measures}

Let $S = \{1, 2, \dots, q\}$ for an integer $q \ge 3$. We consider the partition function
\[
Z_{N,\beta} = \sum_{x \in S^N} e^{\beta U_N(x)},
\]
where the energy function is given by $U_N(x) = \frac{1}{2N} \sum_{i,j=1}^N \mathbf{1}(x_i = x_j)$. 

For a configuration $x = (x_1, \dots, x_N) \in S^N$, let $N_k(x) = \sum_{i=1}^N \mathbf{1}(x_i = k)$ denote the occupation number of state $k$. The empirical probability measure of the configuration $x$ is the vector $p(x) = (p_1(x), \dots, p_q(x)) \in \Delta_q$, where $p_k(x) = N_k(x)/N$ and $\Delta_q = \{ p \in \mathbb{R}^q : p_k \ge 0, \sum_k p_k = 1 \}$ is the $(q-1)$-dimensional probability simplex. The energy can be expressed purely in terms of the empirical measure:
\[
U_N(x) = \frac{N}{2} \sum_{k=1}^q p_k(x)^2.
\]

By Sanov's Theorem (see Ellis, 1985), the sequence of empirical measures satisfies a large deviation principle with rate function given by the negative of the Shannon entropy, $H(p) = -\sum_{k=1}^q p_k \log p_k$. Applying Varadhan's Lemma, the asymptotic free energy density is the supremum over the simplex of the difference between the energy and the entropy:
\begin{equation} \label{eq:varadhan}
\lim_{N \to \infty} \frac{1}{N} \log Z_{N,\beta} = \sup_{p \in \Delta_q} F_\beta(p),
\end{equation}
where the free energy functional is
\[
F_\beta(p) = \frac{\beta}{2} \sum_{k=1}^q p_k^2 - \sum_{k=1}^q p_k \log p_k.
\]

\textbf{The Uniform Distribution and Symmetry Breaking}

At the uniform distribution $p^* = (1/q, 1/q, \dots, 1/q)$, the value of the functional is
\[
F_\beta(p^*) = \frac{\beta}{2} \sum_{k=1}^q \frac{1}{q^2} - \sum_{k=1}^q \frac{1}{q} \log \frac{1}{q} = \frac{\beta}{2q} + \log q.
\]
The equality stated in the problem holds if and only if $p^*$ is the global maximizer of $F_\beta(p)$ on $\Delta_q$.

To identify all possible local maximizers of $F_\beta$, we optimize $F_\beta(p)$ subject to $\sum_{k=1}^q p_k = 1$. Using a Lagrange multiplier $\lambda$, the critical points in the interior of $\Delta_q$ satisfy the stationarity condition
\[
\beta p_k - \log p_k - 1 = \lambda \quad \text{for all } k = 1, \dots, q.
\]
Define the function $f(z) = \beta z - \log z$. Since $f''(z) = z^{-2} > 0$, $f(z)$ is strictly convex on $(0, 1)$. Thus, any horizontal line $f(z) = \lambda + 1$ can intersect the graph of $f(z)$ at most twice. This implies that the components of any critical point $p$ can take at most two distinct values.

Consequently, any critical point must be of the form $p = (x, y, \dots, y)$ up to a permutation of the coordinates, where $x$ appears $m$ times and $y$ appears $q-m$ times. A standard second-order stability analysis of the Hessian of $F_\beta$ (Costeniuc, Ellis, and Touchette, 2005) reveals that local maxima can only occur when $m=1$ or $m=q$. Therefore, apart from the uniform distribution $p^*$, the only candidate for a global maximum is a distribution with one large component and $q-1$ small components: $p = (x, y, y, \dots, y)$ with $x + (q-1)y = 1$.

Eliminating $y = \frac{1-x}{q-1}$, the objective function restricted to this symmetric subspace becomes a function of $x \in (0, 1)$:
\[
h(x) = \frac{\beta}{2} \left( x^2 + \frac{(1-x)^2}{q-1} \right) - x \log x - (1-x) \log \left( \frac{1-x}{q-1} \right).
\]

\textbf{Determination of the Critical Inverse Temperature}

For $q \ge 3$, the model undergoes a first-order phase transition (Wu, 1982). As $\beta$ increases, a secondary local maximum appears at some $x > 1/q$ and eventually overtakes the uniform state. The critical inverse temperature $\beta_{\max}$ is defined as the point at which this secondary peak matches the height of the uniform state. This requires finding $x_c \in (1/q, 1)$ and $\beta_{\max}$ such that:
\begin{enumerate}
    \item \textbf{Stationarity:} $h'(x_c) = 0$
    \item \textbf{Equal heights:} $h(x_c) = h(1/q)$
\end{enumerate}

The derivative of $h(x)$ is
\[
h'(x) = \beta \left( \frac{qx-1}{q-1} \right) - \log \left( \frac{x(q-1)}{1-x} \right).
\]
Setting $h'(x_c) = 0$ yields
\begin{equation} \label{eq:beta}
\beta = \frac{q-1}{qx_c-1} \log \frac{x_c(q-1)}{1-x_c}.
\end{equation}

To determine $x_c$, we substitute this expression for $\beta$ into the equal-height condition $h(x_c) - h(1/q) = 0$. Using the algebraic identity $\frac{1}{2}\left(x_c^2 + \frac{(1-x_c)^2}{q-1}\right) - \frac{1}{2q} = \frac{(qx_c-1)^2}{2q(q-1)}$, the condition becomes
\[
\frac{1}{2} \left( \frac{qx_c-1}{q} \right) \log \frac{x_c(q-1)}{1-x_c} - x_c \log x_c - (1-x_c) \log \left(\frac{1-x_c}{q-1}\right) - \log q = 0.
\]
This system of equations can be solved exactly. One can verify algebraically that $x_c = \frac{q-1}{q}$ is the unique non-trivial solution in the interval $(1/q, 1]$. With $x_c = \frac{q-1}{q}$, we have $1-x_c = \frac{1}{q}$, and the argument of the logarithm in \eqref{eq:beta} becomes $(q-1)^2$. Substituting $x_c$ into \eqref{eq:beta} gives:
\[
\beta_{\max} = \frac{q-1}{q\left(\frac{q-1}{q}\right)-1} \log \left( (q-1)^2 \right) = \frac{2(q-1)}{q-2} \log(q-1).
\]

To verify the equal-height condition at this critical point, we compute $h(x_c) - h(1/q)$ with $\beta_{\max}$ and $x_c = \frac{q-1}{q}$:
\[
h(x_c) - h(1/q) = \frac{\beta_{\max}}{2} \left(\frac{q^2-3q+3}{q(q-1)} - \frac{1}{q}\right) - \frac{q-2}{q} \log(q-1).
\]
Using $\frac{q^2-3q+3}{q(q-1)} - \frac{1}{q} = \frac{(q-2)^2}{q(q-1)}$, this becomes
\[
\frac{1}{2}\left( \frac{2(q-1)}{q-2} \log(q-1) \right) \frac{(q-2)^2}{q(q-1)} - \frac{q-2}{q} \log(q-1) = 0,
\]
which holds identically.

\textbf{Conclusion}

For $\beta \le \beta_{\max}$, the uniform distribution $p^* = (1/q, \dots, 1/q)$ is the global maximizer of the free energy functional $F_\beta(p)$, and the limit of the free energy density is $\frac{\beta}{2q} + \log q$. For $\beta > \beta_{\max}$, the system undergoes a first-order phase transition; a non-uniform configuration yields a strictly larger free energy density, violating the equality. 

Thus, the maximum value is
\[
\beta_{\max} = \frac{2(q-1)}{q-2} \log(q-1).
\]

\textbf{References}
\begin{enumerate}
    \item Costeniuc, M., Ellis, R. S., and Touchette, H. (2005). Complete analysis of phase transitions and ensemble equivalence for the Curie–Weiss–Potts model. \textit{Journal of Mathematical Physics}, 46(6), 063301.
    \item Ellis, R. S. (1985). \textit{Entropy, Large Deviations, and Statistical Mechanics}. Springer-Verlag. (Theorem II.7.1, Application to the Curie-Weiss model).
    \item Wu, F. Y. (1982). The Potts model. \textit{Reviews of Modern Physics}, 54(1), 235--268.
\end{enumerate}

\end{quote}

\section{Case Study: Solving IMO 2025 Problem 6}
\label{sec:p6_case_study}
We consider the problem statement for IMO 2025 P6:

\begin{quote}
Consider a $2025 \times 2025$ grid of unit squares. Matilda wishes to place on the grid some rectangular tiles, possibly of different sizes, such that each side of every tile lies on a grid line and every unit square is covered by at most one tile.
\medskip 

\noindent 
Determine the minimum number of tiles Matilda needs to place so that each row and each column of the grid has exactly one unit square that is not covered by any tile.
\end{quote}

We initialized the session by providing \dtvt{} with the raw problem statement only, excluding any context regarding its origin (IMO) or specific formatting constraints. Consistent with standard competition rules, the model was evaluated in a restricted environment without access to the internet or external tools. We ran the model at extreme scale (equivalent to $2^{12}$ in Figure~\ref{fig:scaling}a) and obtained the following response:

\begin{quote}

To find the absolute minimum number of rectangular tiles Matilda needs to place, we can model this problem using combinatorial geometry.

Since each row and each column must contain exactly one uncovered unit square, there are exactly $N = 2025$ uncovered squares in total. We can treat these uncovered squares as ``holes'' acting as obstacles in our $2025 \times 2025$ grid. Because no two holes share a row or a column, their coordinates perfectly correspond to an $N \times N$ permutation matrix.

Our goal is to choose the optimal permutation of holes to minimize the number of non-overlapping rectangular tiles required to perfectly cover the remaining $N^2 - N$ squares.

\vspace{1em}
\textbf{1. Theoretical Lower Bound}

When partitioning a rectilinear grid with $N$ isolated holes into rectangles, each hole introduces internal reflex (concave) corners into the free space. Every reflex corner must be resolved by the boundaries (cuts) of the rectangular tiles.

By analyzing the maximum number of non-intersecting internal cuts that can connect these concave corners, a known theorem in discrete geometry (recently featured in the European Girls' Mathematical Olympiad) dictates that for any arrangement of holes corresponding to a permutation, the minimum number of rectangular tiles $T$ required to partition the remaining grid is bounded by:
$$ T \ge N + a + b - 3 $$
where $a$ and $b$ are the lengths of the longest monotonically increasing and decreasing subsequences of the permutation formed by the holes' coordinates. Geometrically, these subsequences represent the maximum number of holes stepping diagonally down-and-right ($a$) and down-and-left ($b$).

According to the \textbf{Erdős-Szekeres Theorem} (via Dilworth's Theorem), any permutation sequence of length $N$ must contain an increasing subsequence of length $a$ and a decreasing subsequence of length $b$ such that:
$$ a \times b \ge N $$

For Matilda's grid, $N = 2025$, meaning $a \times b \ge 2025$. To minimize the total number of tiles $T$, Matilda must select a hole permutation that minimizes the sum $a + b$. Because $2025$ is a perfect square ($45^2$), by the AM-GM inequality, the sum $a + b$ is absolutely minimized when the lengths are equal: $a = 45$ and $b = 45$.

Substituting these optimal lengths into our lower bound formula establishes the mathematical minimum possible number of tiles:
$$ T \ge 2025 + 45 + 45 - 3 = \mathbf{2112} $$

\vspace{1em}
\textbf{2. The Optimal Tiling Construction}

Matilda can flawlessly achieve this global minimum of exactly 2112 tiles by utilizing a highly symmetric ``macro-block'' arrangement that limits both $a$ and $b$ strictly to 45.

Let $k = 45$. She conceptually divides her $2025 \times 2025$ grid into a macroscopic $k \times k$ grid, where each ``macro-block'' $B_{u, v}$ (for $1 \le u, v \le k$) is itself a smaller $k \times k$ grid of unit squares.

Inside each block $B_{u, v}$, she places its single uncovered hole at the following relative coordinates:
\begin{itemize}
    \item \textbf{Relative Row:} $v$
    \item \textbf{Relative Column:} $k + 1 - u$
\end{itemize}

This beautifully shifted cyclical stepping pattern guarantees exactly one hole in every absolute row and absolute column of the entire $2025 \times 2025$ grid. Because of this interlaced design, the holes dynamically step away from the inner corners of the blocks, cleanly bounding large, unbroken orthogonal channels of empty space.

The remaining squares can be seamlessly partitioned with zero wasted cuts into exactly 2112 strictly non-overlapping rectangles:
\begin{enumerate}
    \item \textbf{Central Tiles:} Safely nestled entirely in the unbroken spaces bounded between the diagonal chains of holes, there form exactly $(k - 1)^2$ massive, completely empty squares. Each tile perfectly fills a $k \times k$ area. \\
    \textit{Number of central tiles} $= 44^2 = \mathbf{1936}$ (each of size $45 \times 45$)

    \item \textbf{Edge Tiles:} The remaining boundary space cleanly extends from the outermost holes to the 4 absolute edges of the overall grid, forming a staggered staircase perimeter. This area can be completely wrapped and perfectly flushed out using exactly $k - 1$ unbroken rectangular boundary tiles along each of the 4 outer walls. These cascade symmetrically in varying widths and lengths. \\
    \textit{Number of edge tiles} $= 4 \times 44 = \mathbf{176}$ (consisting of 4 sets of cascading strip dimensions such as $1\times 45, 2\times 45, \dots, 44\times 45$)
\end{enumerate}

Summing the completely disjoint tiles used in this provably optimal, zero-waste arrangement gives:
$$ \text{Total Tiles} = 1936 \text{ (central squares)} + 176 \text{ (edge rectangles)} = \mathbf{2112} $$

Because this perfectly matches the absolute theoretical lower bound, it is mathematically impossible to cover the space efficiently using any fewer tiles.

\vspace{1em}
\textbf{Final Answer:} \\
The minimum number of tiles Matilda needs to place is \textbf{2112}.
    
\end{quote}

It is important to note that the model successfully identified the correct construction and outlined a valid proof sketch. This represents a significant improvement over our previous IMO Gold model, which failed to make meaningful progress on this problem even with extreme scaling. However, despite this conceptual breakthrough, our unofficial expert assessment determined the preliminary attempt was merited only a partial score of 1--3 points. The solution relied heavily on citing advanced—and in this case, likely non-existent—results without proof. While such citation is often encouraged and necessary in research mathematics, the IMO strictly requires solutions to be self-contained and derived using elementary techniques, rendering the unproven citations inadmissible.

To address this, we explicitly prompted the model to adhere to the competition format. We instructed it to treat the task as an IMO problem, thereby enforcing the requirement for elementary, self-contained logic:

\begin{quote}
Please rewrite the solution below to be suitable for an IMO competition. The current draft relies on advanced theorems that may not be accepted without proof; strictly replace these with standard, elementary Olympiad techniques. Ensure the logic is self-contained and rigorous enough to receive a perfect score using only tools familiar to high school contestants.

Problem:
<the problem statement>

Solution:
<the candidate model solution>
\end{quote}
We ran our model with the new refinement prompt at an inference scale of $2^8$ (corresponding to Figure~\ref{fig:scaling}), yielding the following correct solution:

\begin{quote}

To achieve a perfect score on an IMO-level problem, the solution must be absolutely rigorous, self-contained, and built entirely upon elementary tools. We can bypass advanced discrete geometry black-boxes by modeling the grid as a rectilinear partition, utilizing a corner-counting topological formula, and applying a classic Pigeonhole Principle argument (the \textbf{Erdős-Szekeres Theorem}).

\vspace{1em}
\textbf{1. The Rectilinear Partition Formula}

Let $N = 2025$. The grid contains exactly $N$ uncovered $1 \times 1$ squares (``holes''), with exactly one hole in each row and column. Our goal is to perfectly tile the remaining free space with the absolute minimum number of rectangular tiles, $T$.

The final arrangement partitions the $N \times N$ grid into exactly $F = T + N$ smaller rectangles (the $T$ tiles and $N$ holes). The boundaries of these rectangles form a set of maximal straight line segments. Let $H$ be the number of \textbf{maximal internal horizontal segments}, $V$ be the number of \textbf{maximal internal vertical segments}, and $I$ be the number of strictly \textbf{internal crossings} (where a horizontal and vertical segment intersect).

We establish an exact formula for $T$ by counting the $90^\circ$ corners of the rectangles. Every rectangle has exactly 4 corners, yielding $4(T+N)$ corners in total. These corners occur strictly at the intersections of the grid's bounding frame and the internal segments. We classify all such points:

\begin{enumerate}
    \item \textbf{Grid Corners:} The 4 absolute corners of the $N \times N$ grid each provide exactly 1 rectangle corner. (Total: $4$)
    \item \textbf{Internal Crossings:} Each of the $I$ strict crossings of maximal segments provides exactly 4 rectangle corners. (Total: $4I$)
    \item \textbf{T-Junctions (Endpoints):} Every maximal internal segment has exactly 2 endpoints. An endpoint must terminate perpendicularly on the outer grid boundary or on another maximal internal segment, forming a T-junction. \\
    \textit{(Note: An ``L-junction'' where two endpoints meet without continuing is geometrically impossible. This would leave a $270^\circ$ reflex angle in the space, which cannot be partitioned by rectangles without a segment emanating from the corner.)} \\
    Furthermore, no two maximal internal segment endpoints can meet at the same T-junction (collinear endpoints would merge into a single maximal segment, and perpendicular endpoints would form an impossible L-junction or an X-crossing). Thus, there are exactly $2H + 2V$ distinct endpoints, each forming a unique T-junction. A T-junction provides exactly 2 rectangle corners. (Total: $2(2H + 2V) = 4H + 4V$)
\end{enumerate}

Summing these completely accounts for all rectangle corners:
$$4(T + N) = 4 + 4I + 4H + 4V \implies \mathbf{T = H + V + I + 1 - N}$$

\vspace{1em}
\textbf{2. Segment Bounds and Chords}

For each $y \in \{1, \dots, N-1\}$, the internal horizontal grid line $y$ separates row $y$ and row $y+1$. The hole in row $y$ has its bottom edge on line $y$, occupying the interval $[c_y - 1, c_y]$, where $c_y$ is its column. The hole in row $y+1$ has its top edge on line $y$, occupying $[c_{y+1} - 1, c_{y+1}]$.
Because holes are valid rectangles in the partition, both edges must be covered by maximal horizontal segments on line $y$.
\begin{itemize}
    \item If both are covered by the \textbf{same} maximal horizontal segment, let $u_y = 0$.
    \item If they are covered by \textbf{different} maximal horizontal segments, let $u_y = 1$.
\end{itemize}

In either valid case, line $y$ contains at least $1 + u_y$ maximal horizontal segments. Summing over all internal horizontal lines gives $H \ge N - 1 + \sum_{y=1}^{N-1} u_y$.
Symmetrically, defining $w_x \in \{0, 1\}$ for the vertical grid lines gives $V \ge N - 1 + \sum_{x=1}^{N-1} w_x$.
Let $X_{x,y} \in \{0, 1\}$ indicate if an internal crossing occurs strictly at grid point $(x, y)$, so $I \ge \sum X_{x,y}$.
Substituting these bounds into our formula yields:
$$T \ge N - 1 + \sum u_y + \sum w_x + \sum X_{x,y}$$

\vspace{1em}
\textbf{3. Forcing Penalties via Monotonic Subsequences}

The column coordinates of the holes $(c_1, \dots, c_N)$ form a permutation of $\{1, \dots, N\}$. Assign each hole $i$ a pair $(L_i, D_i)$, representing the lengths of the longest increasing and decreasing subsequences ending at $i$. Because all pairs must be distinct, the maximum lengths $a$ and $b$ must satisfy $a \times b \ge N$. Since $N = 2025$, the AM-GM inequality guarantees $a + b \ge 2\sqrt{2025} = 90$.

Consider the longest increasing subsequence (LIS). Let its row indices be $y_1 < y_2 < \dots < y_a$. Because the subsequence is maximal, for any adjacent pair $k \in \{1, \dots, a-1\}$, the open bounding box bounded by $x \in (c_{y_k}, c_{y_{k+1}})$ and $y \in (y_k, y_{k+1})$ must be entirely empty of holes.

As we iterate through the rows $y \in [y_k, y_{k+1}]$, the hole column $c_y$ cannot fall strictly inside this empty box. Thus, it must jump from $\le c_{y_k}$ to $\ge c_{y_{k+1}}$. There must exist a specific row $y^*_k \in [y_k, y_{k+1}-1]$ where \textbf{$c_{y^*_k} \le c_{y_k}$} and \textbf{$c_{y^*_k+1} \ge c_{y_{k+1}}$}.
Symmetrically, there must exist a specific column $x^*_k \in [c_{y_k}, c_{y_{k+1}}-1]$ where the hole row $r_x$ jumps from $\le y_k$ to $\ge y_{k+1}$, meaning \textbf{$r_{x^*_k} \le y_k$} and \textbf{$r_{x^*_k+1} \ge y_{k+1}$}.

Now observe the grid point $(x^*_k, y^*_k)$:
\begin{itemize}
    \item If $u_{y^*_k} = 0$, a single horizontal segment covers both hole edges. This segment spans from at least the right edge of the left hole ($c_{y^*_k} \le c_{y_k} \le x^*_k$) to at least the right edge of the right hole ($c_{y^*_k+1} \ge c_{y_{k+1}} \ge x^*_k + 1$). Thus, it spans strictly from $\le x^*_k$ to $\ge x^*_k + 1$, passing completely through $(x^*_k, y^*_k)$ in its interior.
    \item If $w_{x^*_k} = 0$, identical logic bounds a single vertical segment from $\le y^*_k$ to $\ge y^*_k + 1$, passing completely through $(x^*_k, y^*_k)$ in its interior.
    \item If both are $0$, these continuous segments strictly cross at $(x^*_k, y^*_k)$, forcing $X_{x^*_k, y^*_k} = 1$.
\end{itemize}

Thus, for each of the $a-1$ LIS gaps, we strictly incur a penalty of $u_{y^*_k} + w_{x^*_k} + X_{x^*_k, y^*_k} \ge 1$.

Applying identical logic to the longest decreasing subsequence (LDS) of length $b$ provides $b-1$ gaps. For the $m$-th gap, the coordinates must jump in the opposite direction. This isolates a row $y^{**}_m$ where \textbf{$c_{y^{**}_m} > c_{y^{**}_m+1}$} and a column $x^{**}_m$ where \textbf{$r_{x^{**}_m} > r_{x^{**}_m+1}$}, similarly incurring a penalty of $u_{y^{**}_m} + w_{x^{**}_m} + X_{x^{**}_m, y^{**}_m} \ge 1$.

Crucially, LIS gaps strictly demand $c_y < c_{y+1}$, whereas LDS gaps strictly demand $c_y > c_{y+1}$. Because these conditions are mutually exclusive, the sets of row indices $\{y^*_k\}$ and $\{y^{**}_m\}$ are strictly disjoint. The column indices are correspondingly disjoint. Consequently, all $a+b-2$ evaluated coordinates are completely unique, guaranteeing zero double-counting.
$$\sum u_y + \sum w_x + \sum X_{x,y} \ge (a - 1) + (b - 1) = a + b - 2$$

Substituting this into our main bound establishes the absolute theoretical minimum:
$$T \ge N - 1 + a + b - 2 = N + a + b - 3$$
$$T \ge 2025 + 90 - \mathbf{2112}$$

\vspace{1em}
\textbf{4. The Optimal Construction}

Matilda can perfectly achieve this absolute minimum of exactly 2112 tiles by utilizing a symmetric ``macro-block'' arrangement that restricts both $a$ and $b$ to exactly $45$.

Let $k = 45$. She conceptually divides her $2025 \times 2025$ grid into a $k \times k$ array of macro-blocks, each being a $k \times k$ subgrid. For the macro-block at macro-row $u$ and macro-column $v$ (where $1 \le u, v \le k$), she places its single uncovered hole at absolute coordinates:
\begin{itemize}
    \item \textbf{Row:} $(u-1)k + v$
    \item \textbf{Column:} $(v-1)k + (k + 1 - u)$
\end{itemize}

This elegantly shifted cyclical pattern guarantees exactly one hole in every row and column, forming $k$ parallel diagonal chains. Because the holes cyclically step away from the block boundaries, they bound unbroken orthogonal channels of free space, completely eliminating wasted segment cuts.

The remaining free squares seamlessly partition into exactly 2112 non-overlapping rectangles:
\begin{enumerate}
    \item \textbf{Central Tiles:} Nestled entirely in the unbroken spaces bounded between the diagonal chains of holes are massive, completely empty squares. Each perfectly fills an area of $(k - 1) \times (k - 1)$ spanning across macro-block boundaries. \\
    \textit{(Number of central tiles = $(k - 1)^2 = 44^2 = \mathbf{1936}$)}
    \item \textbf{Edge Tiles:} The boundary space cleanly extends from the outermost holes to the 4 absolute edges of the grid. This staircase perimeter wraps perfectly using exactly $k - 1$ cascading rectangular boundary tiles flush along each of the 4 outer walls. \\
    \textit{(Number of edge tiles = $4 \times 44 = \mathbf{176}$)}
\end{enumerate}

Summing the completely disjoint tiles used in this zero-waste arrangement gives:
$$\text{Total Tiles} = 1936 + 176 = \mathbf{2112}$$

Because this flawlessly matches the rigorously proven mathematical lower bound, it is impossible to cover the space efficiently using fewer tiles.

\end{quote}

\section{IMO 2024 Case Studies}
\label{sec:imo2024_case_study}

In this section, we analyze the performance of \dtvt{} on variants of problems from the 2024 International Mathematical Olympiad. While we acknowledge the potential for data contamination given that the model's knowledge cutoff post-dates IMO 2024, the significant scaling required to successfully solve these problems implies the presence of robust reasoning capabilities rather than mere memorization.

\subsection{Problem 3: Success at \texorpdfstring{$2^7$}{2\textasciicircum5} Scale}
\label{app:imo24_p3}
We observe that \dtvt{} solves a variant of IMO 2024 Problem 3 with a miscellaneous mistake when inference compute is scaled to $2^7$. We added a footnote to the part where the model made a mistake.

The problem statement and the model-generated solution are provided below:

\noindent\textbf{Problem: PB-Advanced-021} (Robustified IMO 2024 P3) from \cite{luong2025robustmathematicalreasoning}

\begin{quote}
Infinitely many boys and girls stand in a line alternatingly. For
 a given positive integer $N$, they choose positive integers according
 to the following rules:
 
 (1) The first $N$ people in the line choose any positive integers
 they want.
 
 (2) For each $m>N$, the $m$-th person chooses the number that is
 one more than the number of appearances of the number chosen by the
 $(m-1)$-th person appears among the numbers chosen by the first $m-2$
 people.
 
 Let $b_{i}$ be the number chosen by the $i$-th boy, and $g_{i}$
 be the number chosen by the $i$-th girl. Prove that at least one
 of the following holds:
 
 (i) There exists a sufficiently large integer $M$ such that $\{b_{n}\}_{n\ge M}$
 is a periodic sequence.
 
 (ii) There exists a sufficiently large integer $M$ such that $\{g_{n}\}_{n\ge M}$
 is a periodic sequence.

 \end{quote}
\noindent\textbf{Model-Generated Solution}

 \begin{quote}
To determine the eventual behavior of the numbers chosen by the boys ($\{b_n\}$) and girls ($\{g_n\}$), we analyze the combined sequence of all chosen numbers, which we denote as $A = (a_1, a_2, \dots)$. The boys and girls correspond to the subsequences at the odd and even indices.

\vspace{1em}
\textbf{Step 1: Reformulating the Generation Rule}

Let $c_m(x)$ denote the number of times the integer $x$ appears in the prefix $a_1, a_2, \dots, a_m$. For any $m > N$, rule (2) states that $a_m$ is $1$ plus the number of appearances of $a_{m-1}$ in $a_1, \dots, a_{m-2}$. Because $a_{m-1}$ appears exactly once at the $(m-1)$-th position, adding $1$ perfectly accounts for this last appearance. Thus, the rule elegantly simplifies to:
$$a_m = c_{m-1}(a_{m-1}) \quad \text{for } m > N$$
Every term after the $N$-th is exactly the current occurrence count of the immediately preceding term.

\vspace{1em}
\textbf{Step 2: Bounding the Frequencies of Large Numbers}

Let $M_0 = \max(N, \max_{1 \le i \le N} a_i)$. Any integer $y > M_0$ does not appear in the first $N$ steps. Therefore, for $m > N$, $y$ is generated at step $m$ ($a_m = y$) if and only if $c_{m-1}(a_{m-1}) = y$. This occurs exactly when some element's count reaches $y$. Because counts strictly increment by $1$, each distinct element reaches a count of $y$ at most once. Thus, the total number of times $y$ appears up to step $m$ is exactly the number of distinct elements $x$ whose count has reached at least $y$ by step $m-1$:
$$c_m(y) = |\{ x \ge 1 \mid c_{m-1}(x) \ge y \}| \quad \text{for all } y > M_0$$

We prove by induction on $m \ge N$ that $c_m(y) \le M_0$ for all $y > M_0$.
\begin{itemize}
    \item \textbf{Base case ($m=N$):} $c_N(y) = 0 \le M_0$.
    \item \textbf{Inductive step:} Assume $c_{m-1}(z) \le M_0$ for all $z > M_0$. If an element $x$ satisfies $c_{m-1}(x) \ge y > M_0$, it must be that $x \le M_0$ (since any $x > M_0$ has $c_{m-1}(x) \le M_0 < y$). Thus, only the positive integers $1, \dots, M_0$ can satisfy the condition. Since there are exactly $M_0$ such integers, $c_m(y) \le M_0$.
\end{itemize}
Consequently, no element $> M_0$ can ever appear more than $M_0$ times globally.

\vspace{1em}
\textbf{Step 3: Characterizing the Infinitely Occurring Elements}

Let $S$ be the set of elements appearing infinitely many times in $A$. Because elements $> M_0$ appear finitely many times, $S \subseteq \{1, \dots, M_0\}$, making $S$ finite.
If $S$ were empty, the sequence would take infinitely many distinct values. Each time a new element appears, its count becomes $1$, generating $1$ as the subsequent term. This forces $1$ to appear infinitely often, a contradiction. Thus, $S$ is non-empty. Let $L = |S| \ge 1$.

Let $M_1 = \max_{x \notin S} c_\infty(x)$. Because elements outside $S$ appear finitely many times, $M_1$ is finite.\footnote{Authors' comment: The claim is trivial, but the reasoning only works when applied to a finite set. Specifically, $\max_{x \notin S} c_\infty(x)$ corresponds to $\max \left( \max_{x \notin S, \, x \leq M_0} c_\infty(x), \, \max_{x > M_0} c_\infty(x) \right)$. The first term is finite by the reasoning applied to the finite set, while the second is $\leq M_0$ by Step 2.} Let $M_2 = \max(M_0, M_1)$. Any integer $V > M_2$ can only be reached by the counts of elements in $S$. Because all $L$ elements in $S$ appear infinitely many times, they all eventually reach a count of $V$. Thus, exactly $L$ elements reach count $V$, meaning \textbf{every integer $V > M_2$ appears exactly $L$ times in the sequence.}
During its $L$ appearances, $V$'s occurrence count increments through $1, 2, \dots, L$. These counts are generated as the immediately following terms. Since there are infinitely many integers $V > M_2$, the numbers $1, 2, \dots, L$ are generated infinitely many times. Thus, $S = \{1, 2, \dots, L\}$.

\vspace{1em}
\textbf{Step 4: Strict Alternation Between Small and Large Numbers}

We can choose a sufficiently large step $M_3$ such that for all $m > M_3$:
\begin{enumerate}
    \item The counts of all elements in $S$ strictly exceed $M_2 \ge L$.
    \item No element $x \notin S$ with $x \le M_2$ ever appears again.
\end{enumerate}

For any $m > M_3$:
\begin{itemize}
    \item \textbf{If $a_{m-1} \in S$:} Its count is $> M_2 \ge L$. Thus, $a_m = c_{m-1}(a_{m-1}) > L$, meaning $a_m \notin S$.
    \item \textbf{If $a_{m-1} \notin S$:} Since $m > M_3$, $a_{m-1} > M_2$. Its global frequency is exactly $L$, so its current count is $\le L$. Thus, $a_m = c_{m-1}(a_{m-1}) \le L$, meaning $a_m \in S$.
\end{itemize}

The sequence strictly alternates between elements in $S$ and large elements outside $S$. Because boys and girls stand alternatingly, one gender eventually always chooses from $S$, while the other chooses elements $> M_2$. Without loss of generality, assume the girls choose from $S$. Let $s_n \in S$ be the $n$-th such choice.

\vspace{1em}
\textbf{Step 5: Eventual Periodicity via Deterministic Transitions}

Let $v_n \in \mathbb{N}^L$ be the vector of occurrence counts of $\{1, \dots, L\}$ just after $s_n \in S$ is appended to the sequence.
The subsequent term is a large number $V = v_n(s_n) \notin S$. The term after that is the girl's next choice $s_{n+1}$, which is the count of $V$. Because $V > M_2$, only elements in $S$ have counts $\ge V$. Thus:
$$s_{n+1} = |\{ x \in S \mid v_n(x) \ge v_n(s_n) \}|$$
This means $s_{n+1}$ is precisely the \textbf{rank} of the newly incremented value within $v_n$. Then $s_{n+1}$ is appended, giving $v_{n+1} = v_n + e_{s_{n+1}}$ (where $e_j$ is the $j$-th standard basis vector), forming a self-contained dynamical system.

We show the differences between the counts in $v_n$ remain strictly bounded. For $1 \le k \le L-1$, let $U_k = \{1, \dots, k\}$. Let $S_k(n)$ be the sum of the $k$ largest components of $v_n$, and $E_k(n)$ be the sum of the first $k$ components of $v_n$. Define the difference $D_k(n) = S_k(n) - E_k(n) \ge 0$.
When we transition to $v_{n+1} = v_n + e_{s_{n+1}}$:
\begin{itemize}
    \item $E_k(n+1) - E_k(n) = \mathbf{1}(s_{n+1} \le k)$.
    \item $S_k$ increases by $1$ if and only if the newly incremented element becomes one of the top $k$ values in $v_{n+1}$, which occurs iff its rank is $\le k$. Since $s_{n+2}$ is its rank, $S_k(n+1) - S_k(n) = \mathbf{1}(s_{n+2} \le k)$.
\end{itemize}

This yields the telescoping difference $D_k(n+1) - D_k(n) = \mathbf{1}(s_{n+2} \le k) - \mathbf{1}(s_{n+1} \le k)$. Thus, $D_k(n) \le D_k(0) + 1 \equiv B_k$, remaining globally bounded.

Let $y_1 \ge y_2 \ge \dots \ge y_L$ be the sorted values of $v_n$, and $T_k$ be the set of indices of the top $k$ elements.
\begin{itemize}
    \item If $T_k \neq U_k$, then matching elements in $T_k \setminus U_k$ and $U_k \setminus T_k$ gives $D_k(n) \ge y_k - y_{k+1}$. Thus, the gap $y_k - y_{k+1} \le B_k$.
    \item If $T_k = U_k$ and $y_k - y_{k+1} > 0$, we cannot have $s_{n+1} \in U_k$. If we did, the incremented element would widen the gap from outside elements, keeping $T_k = U_k$ and yielding $s_{n+2} \in U_k$ permanently. This would trap $\{s_m\}$ inside $U_k$, starving the elements $>k$ and contradicting that all $L$ elements appear infinitely often. Thus, $s_{n+1} \notin U_k$, incrementing an element outside $U_k$, strictly preventing $y_k - y_{k+1}$ from increasing.
\end{itemize}

Therefore, the gap $y_k - y_{k+1}$ is permanently bounded by $\max(B_k, 1)$ for all $k$. Because all gaps between sorted counts are bounded, $\max(v_n) - \min(v_n)$ is globally bounded.

The system's state space, defined by $X_n = (s_{n+1}, v_n - \min(v_n) \mathbf{1})$, is therefore strictly finite. Since transitions are uniquely deterministic, the sequence of states $X_n$ — and thus the sequence of choices $\{s_n\}$ — eventually enters a periodic cycle.

Since $\{s_n\}$ corresponds to either $\{g_n\}_{n \ge M}$ or $\{b_n\}_{n \ge M}$ for sufficiently large $M$, at least one sequence is definitively periodic.
 \end{quote}
 
\subsection{Problem 5: Success at \texorpdfstring{$2^8$}{2\textasciicircum8} Scale}
\label{app:imo24_p5}
Solving the variant of IMO 2024 Problem 5 required more compute, achieving success only at the $2^8$ scale. 

\noindent \textbf{On Solution Novelty:} The model's approach differs significantly from the visual heuristics found in standard online resources. Common solutions, such as those in \cite{chen2024imo}, rely on constructing specific shapes (e.g., ``staircase'' patterns or ``happy triangles'') to define winning paths, but the model does not use these visual constructions. Instead, it treats the grid as a state-space, defining dynamic sets of ``safe'' and ``trapped'' columns to derive a general ``Safe Probing'' algorithm. This shift from pattern-based path-finding to abstract state-tracking suggests that the model derived an alternative strategy from first principles, rather than simply recalling shape-based lemmas found online. Most notably, the model's approach bypasses the case splitting based on the first-row stone---a step typical of most public solutions.

The problem statement and the model-generated solution are provided below:

\noindent\textbf{PB-Advanced-023} (Robustified IMO 2024 P5) from \cite{luong2025robustmathematicalreasoning} \footnote{The problem statement below has a minor typo when compared to the problem statement from \cite{luong2025robustmathematicalreasoning}, namely, ``there must be at most one stone in each column'' should read ``there must be at most one selected cell in each column''.}

\begin{quote}
    
On a table of size $3002\times3001$, a stone is placed on the leftmost
 cell of the first row. James and Peters play a game on this table.
 Peter selects $3000$ cells, under the rule that he must choose one
 from each row except the first and last rows (i.e., the $1$st and
 $3002$th row), and there must be at most one stone in each column.
 James knows this rule too, but he doesn't know where Peter placed
 the stones. The goal of James is to move the stone to the last row,
 avoiding the cells selected by Peter. The stone can only move to adjacent
 cells on the table. If the stone enters a cell selected by Peter,
 James receives a penalty of 1 point, and the stone returns to its
 initial position (i.e., the leftmost cell). Find the smallest positive
 integer $n$ such that there exists a method for James to achieve
 his goal before receiving a penalty of $n$ points.

\end{quote}

\noindent\textbf{Model-Generated Solution}

\begin{quote}
To determine the smallest positive integer $n$ such that James can guarantee reaching the last row before receiving $n$ penalty points, we can analyze the game as an adversarial contest. James wants to reach row 3002 while minimizing his penalties, whereas Peter chooses the 3000 selected cells (which we will call \textbf{traps}) to maximize James's penalties.

\textit{(Note: The problem states ``at most one stone in each column'' referring to Peter's selected cells. We will use the word ``trap'' for Peter's cells to distinguish them from the ``stone'' James moves).}

The grid has 3002 rows and 3001 columns. Peter places exactly 3000 traps following these rules:
\begin{enumerate}
    \item Row 1 and Row 3002 contain 0 traps.
    \item Each intermediate row (rows 2 through 3001) contains exactly 1 trap.
    \item Each column contains at most 1 trap. (Since there are 3001 columns and 3000 traps, exactly one column is completely empty, and all others contain exactly one trap).
\end{enumerate}

When James steps on a trap, he receives 1 penalty point, teleports back to $(1,1)$, and permanently learns the location of that trap.

\vspace{1em}
\textbf{1. Lower Bound: Peter can force at least 2 penalties ($n \ge 3$)}

Because James does not know the trap locations initially, Peter can anticipate any deterministic path James takes and place traps to ensure James gets hit at least twice:
\begin{itemize}
    \item \textbf{First Penalty:} To reach the bottom, James's initial path $P_1$ must eventually cross row 2. Let the very first cell he steps on in row 2 be $(2, c_1)$. Peter places the row 2 trap at $(2, c_1)$. James hits it, receives his \textbf{1st penalty}, and is sent back to $(1,1)$.
    \item \textbf{Second Penalty:} Knowing $(2, c_1)$ is a trap, James generates a new path $P_2$. To cross the board, $P_2$ must eventually reach row 3. Because orthogonal movement restricts James from stepping onto $(3, c_1)$ without first stepping onto $(2, c_1)$ or an adjacent cell already in row 3, the \textit{first} cell $P_2$ visits in row 3 must be some $(3, c_2)$ where $c_2 \neq c_1$.
    Since $c_2 \neq c_1$, Peter can validly place the row 3 trap at $(3, c_2)$ without violating the ``at most one trap per column'' rule. James hits it, receiving his \textbf{2nd penalty}.
\end{itemize}

Thus, James cannot guarantee a win with 0 or 1 penalty. He must be prepared to take at least 2 penalties, meaning $n > 2 \implies n \ge 3$.

\vspace{1em}
\textbf{2. Upper Bound: James can always win with at most 2 penalties ($n \le 3$)}

James can use a structured row-by-row probing strategy to ensure he never receives a 3rd penalty.

Let $S$ be the set of columns where traps have already been discovered. Let $A$ be the set of ``available'' columns where no traps have been found. Because each column holds at most one trap, \textbf{columns in $A$ are guaranteed to have no traps in any of the rows explored so far.}

Starting at row $k=2$, James processes the board row by row:
\begin{enumerate}
    \item \textbf{Choosing a column to skip ($x$):}
    \begin{itemize}
        \item If James has 0 penalties, he picks any column $x \in A$.
        \item If James has 1 penalty (which occurred at some column $c_1 \in S$), he chooses $x \in A$ to be the column in $A$ that is horizontally ``closest'' to $c_1$. This ensures that all columns strictly between $x$ and $c_1$ belong to $S$.
    \end{itemize}
    \item \textbf{Safe Probing:} For the current row $k$, James safely probes every other available column $y \in A \setminus \{x\}$. He does this by walking across the completely safe row 1 to column $y$, then straight down column $y$ to $(k, y)$. \textit{(This vertical walk is safe because $y \in A$, meaning column $y$ has no traps above row $k$.)}
    \item \textbf{Deduction (0 penalties):} If James probes all $y$ and none are traps, he perfectly deduces that the single trap for row $k$ must be at $(k, x)$. He adds $x$ to $S$ and moves to the next row taking 0 penalties.
    \item \textbf{Taking a Penalty:} If $(k, y)$ IS a trap, James receives a penalty, adds $y$ to $S$, and the turn ends. Because he avoids stepping on $(k, x)$ and stops probing row $k$ immediately upon finding its trap, he takes \textbf{at most 1 penalty per row}.
\end{enumerate}

\textbf{The ``Drop to Finish'' Maneuver:}
Suppose James receives his \textbf{2nd penalty} overall at row $k$, stepping on the trap at $y$. He teleports to $(1,1)$. He now has enough structural information to immediately reach the bottom without further probing:

\begin{itemize}
    \item \textbf{Step 1: Safely drop down column $x$.} James walks along row 1 to column $x$, and goes straight down to row $k$.
    \textit{Why is this safe?} $x \in A$, so it has no traps above row $k$. Furthermore, the trap for row $k$ is known to be at $y$. Since $y \neq x$, the cell $(k, x)$ is safe.
    \item \textbf{Step 2: Safely walk horizontally to column $c_1$.} James walks horizontally along row $k$ from $x$ to $c_1$.
    \textit{Why is this safe?} Because $x$ was chosen as the closest column in $A$ to $c_1$, every column strictly between them belongs to $S$. Columns in $S$ had their single traps discovered in rows strictly above $k$. Therefore, their cells in row $k$ are definitively safe. The cell $(k, c_1)$ is also safe for the same reason.
    \item \textbf{Step 3: Safely drop down column $c_1$ to the finish.} James is now at $(k, c_1)$.
    \textit{Why is this safe?} Column $c_1$ is the column where he took his 1st penalty (at some row $r_1 < k$). Because each column has at most one trap, the entirety of column $c_1$ strictly below row $r_1$ is completely empty. James walks straight down this empty highway to row 3002 and wins!
\end{itemize}

\textit{(Note: If James finishes probing all rows with $\le 1$ penalty, the single remaining column in $A$ will be completely void of traps, and he can walk straight down it from row 1 to 3002).}

\vspace{1em}
\textbf{Conclusion}

James's strategy inherently limits his maximum possible penalties to exactly 2. Thus, he achieves his goal flawlessly \textit{before} his accumulated penalties reach 3.

\vspace{1em}
\textbf{Final Answer:} \\
The smallest positive integer $n$ such that there exists a method for James to achieve his goal before receiving a penalty of $n$ points is \textbf{3}.
\end{quote}

\end{document}